\documentclass[letterpaper]{article} % DO NOT CHANGE THIS
\usepackage{aaai25}  % DO NOT CHANGE THIS
\usepackage{times}  % DO NOT CHANGE THIS
\usepackage{helvet}  % DO NOT CHANGE THIS
\usepackage{courier}  % DO NOT CHANGE THIS
\usepackage[hyphens]{url}  % DO NOT CHANGE THIS
\usepackage{graphicx} % DO NOT CHANGE THIS
\usepackage{natbib}  % DO NOT CHANGE THIS AND DO NOT ADD ANY OPTIONS TO IT
\usepackage{caption} % DO NOT CHANGE THIS AND DO NOT ADD ANY OPTIONS TO IT
\usepackage{algorithm}
\usepackage{algorithmic}
\usepackage{amsmath}
\usepackage{adjustbox}
\usepackage{multicol}
\usepackage{multirow}
\usepackage{appendix}
\usepackage{marvosym}

\usepackage{amsfonts}
\usepackage{CJKutf8}
\usepackage{newfloat}
\usepackage{listings}
\DeclareCaptionStyle{ruled}{labelfont=normalfont,labelsep=colon,strut=off} % DO NOT CHANGE THIS
\floatstyle{ruled}
\newfloat{listing}{tb}{lst}{}
\floatname{listing}{Listing}
\title{EvdCLIP: Improving Vision-Language Retrieval with Entity Visual Descriptions from Large Language Models}
\author{
    GuangHao Meng\textsuperscript{\rm 1,2}, Sunan He\textsuperscript{\rm 6}, Jinpeng Wang\textsuperscript{\rm 1}, Tao Dai\textsuperscript{\rm 4}, Letian Zhang\textsuperscript{\rm 1}, Jieming Zhu\textsuperscript{\rm 5}\thanks{Project lead}, \\Qing Li\textsuperscript{\rm 2}, Gang Wang\textsuperscript{\rm 5}, Rui Zhang\textsuperscript{\rm 3 \Letter}, Yong Jiang\textsuperscript{\rm 1,2 \Letter}\thanks{\Letter~Corresponding authors.}
}
\affiliations{
    \textsuperscript{\rm 1}Tsinghua Shenzhen International Graduate School, Tsinghua University\\
    \textsuperscript{\rm 2}Peng Cheng Laboratory\\
    \textsuperscript{\rm 3}School of Computer Science \& Tech, Huazhong University of Science and Technology\\
    \textsuperscript{\rm 4}College of Computer Science and Software Engineering, Shenzhen University\\
    \textsuperscript{\rm 5}Huawei Noah’s Ark Lab \\
    \textsuperscript{\rm 6}Hong Kong University of Science and Technology\\
    \{menggh22, wjp20, zlt23\}@mails.tsinghua.edu.cn, daitao.edu@gmail.com, liq@pcl.ac.cn \\ 
    jiemingzhu@ieee.org, \textsuperscript{\rm 3}rayteam@yeah.net, jiangy@sz.tsinghua.edu.cn\\
}

\usepackage{bibentry}
\begin{document}

\maketitle

\begin{abstract}
Vision-language retrieval (VLR) has attracted significant attention in both academia and industry, which involves using text (or images) as queries to retrieve corresponding images (or text). However, existing methods often neglect the rich visual semantics knowledge of entities, thus leading to incorrect retrieval results. To address this problem, we propose the Entity Visual Description enhanced CLIP (EvdCLIP), designed to leverage the visual knowledge of entities to enrich queries. Specifically, since humans recognize entities through visual cues, we employ a large language model (LLM) to generate Entity Visual Descriptions (EVDs) as alignment cues to complement textual data. These EVDs are then integrated into raw queries to create visually-rich, EVD-enhanced queries. Furthermore, recognizing that EVD-enhanced queries may introduce noise or low-quality expansions, we develop a novel, trainable EVD-aware Rewriter (EaRW) for vision-language retrieval tasks. EaRW utilizes EVD knowledge and the generative capabilities of the language model to effectively rewrite queries. With our specialized training strategy, EaRW can generate high-quality and low-noise EVD-enhanced queries. Extensive quantitative and qualitative experiments on image-text retrieval benchmarks validate the superiority of EvdCLIP on vision-language retrieval tasks.
\end{abstract}

\section{Introduction}

Vision-language retrieval (VLR), as a fundamental task in vision-language applications, has attracted extensive research and industrial interest due to its significant research and practical value. It usually takes descriptive texts as queries and retrieves corresponding images, or vice versa. 
\par
\begin{figure}
  \centering
  \includegraphics[width=1.0\linewidth]{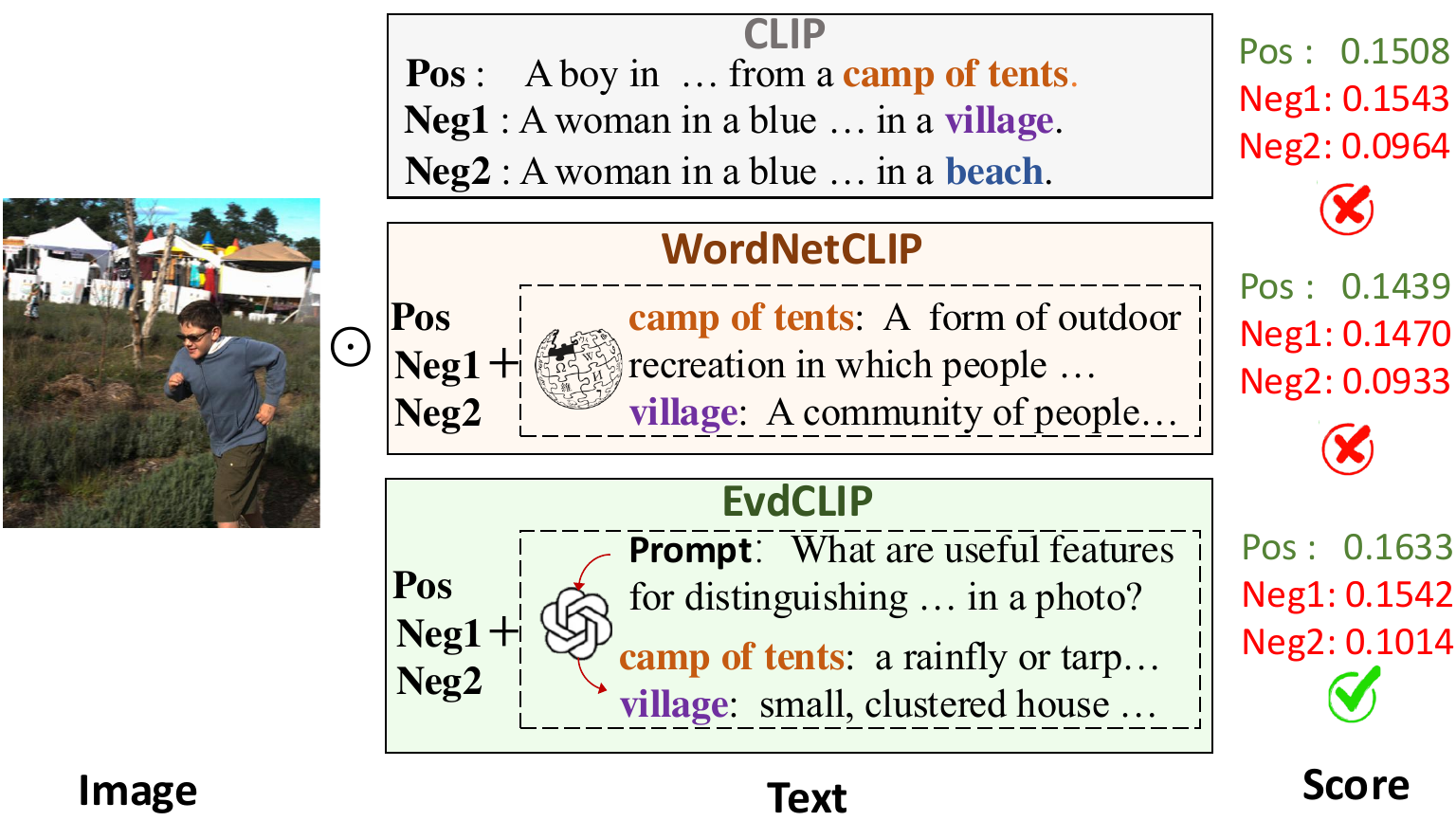}
  \caption{Illustration of entity visual descriptions (EVD) enhanced framework. The CLIP and WordNetCLIP which introduces the concept of entities struggle to distinguish between "camping of tents" and "village", leading to incorrect retrieval results. Our EvdCLIP leverages the  EVD generated by LLMs to improve cross-modal retrieval performance.}
  \label{fig:teaser}
\end{figure}

Existing methods heavily rely on the alignment between visual and textual representations. As shown in Figure~\ref{fig:teaser}, CLIP~\cite{radford2021learning} successfully differentiates between "beach" and "camp of tents" but confuses "camp of tents" with "village", leading to incorrect retrievals. Even with descriptions from WordNet~\cite{kilgarriff2000wordnet}, it struggles to distinguish these concepts. We argue that the lack of visual information in these descriptions and that visual descriptions are crucial to distinguish visually similar entities.

\par
Let's start by analyzing how humans recognize entities in an image. Humans are able to easily describe the visual features of entities using language and leverage these visual descriptions to enhance perception, even for unfamiliar entities. Our key insights are: (1) Visual descriptions offer textual additional cues that improve image-text alignment. (2) Descriptions highlight critical details and discriminative information, aiding in entity recognition.
(3) They encompass generic features, boosting the model's transferability.

\par
However, existing methods struggle to obtain EVD to offer useful cues in multi-modal retrieval. Manually creating these descriptions is costly and impractical given the vast number of concepts in our world. Recently, with the advance in Large Language Models (LLMs), several works utilize the LLMs to generate training samples or auxiliary information for specific tasks~\cite{touvron2023llama, zhu2023minigpt, liu2023llava}. The large-scale corpus used to train these LLMs contains a substantial amount of semantic knowledge, making them into rich visual knowledge bases.
% However, existing multi-modal retrieval (MMR) models have not explored how to leverage the powerful knowledge of LLMs to facilitate the multi-modal retrieval task.
\par
Based on these insights, we propose Entity Visual Descriptions enhanced CLIP (EvdCLIP), which leverages LLMs to generate valuable visual descriptions as auxiliary cues to guide VLR. Specifically, we first employ LLMs to create an Entity Visual Descriptions (EVD) knowledge base from the image-text dataset. Subsequently, EVD knowlege  base is then used to enhance queries with visual descriptions, enabling cross-modal alignment between text and images.

\begin{figure}[t]
\centering
\includegraphics[width=0.85\linewidth]{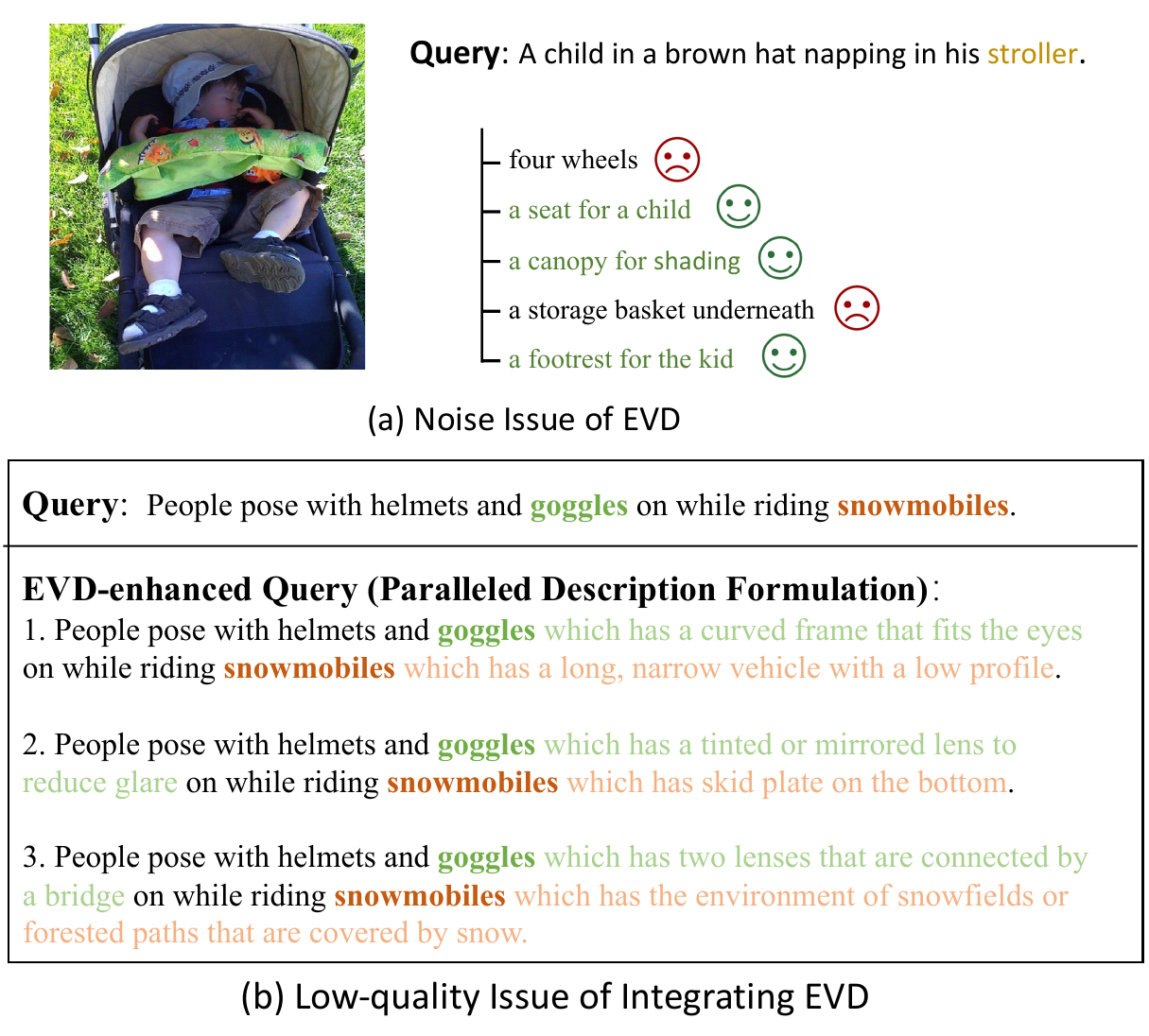}
\caption{Challenges of EVD integration to VLR. (a) Noise issue. Certain descriptions (e.g., "four wheels") may not be presented in the "stroller" in the image and query helps to reveal the entity's preferences. (b) Low-quality issue. Using templates "which has/is" to concatenate entities and descriptions can compromise fluency and introduce ambiguity.}
\label{fig:noise_sample}
\end{figure}

Although some research \cite{yao2022detclip,menon2022visual,maniparambil2023enhancing,yang2023language,pratt2023does,an2023more} has applied descriptions to image classification and object detection, considering that queries in VLR are complex sentences containing multiple entities, applying EVDs to VLR presents two challenges: noise and low-quality issue. The noise issue arises because EVDs exhibit over-diversity due to the lack of constraints specific to the image, leading to inconsistencies in some EVDs. As shown in Fig~\ref{fig:noise_sample} (a), we should consider query content for EVD's denoising. As illustrated in Fig~\ref{fig:noise_sample} (b), the low-quality issue occurs when existing parallel description paradigm's combining query and description leads to awkward and unsmooth queries.

To address these challenges, we introduce an EVD-aware rewriter (EaRw) that dynamically selects EVDs based on the query, generating high-quality VLR queries. To bridge the gap between knowledge-enhanced tasks and pre-trained rewriters, we create a trainable scheme. Using LLM's ability and CLIP's feedback, we generate a high-quality corpus that captures context preferences and dataset preferences~\cite{dunlap2024describing}. EaRw then learns to effectively select and integrate EVDs based on query, mitigating noise and low-quality issues, and enhancing VLR performance.

\par
The contributions of our work are three-fold: \textit{\textbf{(1)}} We propose \textbf{EvdCLIP}, utilizing LLM-based visual descriptions to improve visual-linguistic alignment in VLR. To our knowledge, this is the pioneering effort to use LLMs' visual knowledge for guiding VLR. \textit{\textbf{(2)}} We develop a novel \textbf{EVD-aware Rewriter (EaRW)} using the compact, trainable T5~\cite{raffel2020exploring} model to generate precise and fluent EVD-enhanced queries, effectively mitigating noise of EVD and enhancing query quality. \textit{\textbf{(3)}} We conduct extensive experiments to \textbf{validate the effectiveness of our method on the public benchmark and Huawei business data}. In addition, we highlight the benefits of our framework in terms of generalizability, transferability, and editability.

\begin{figure*}
  \centering
  \includegraphics[width=0.8\linewidth]{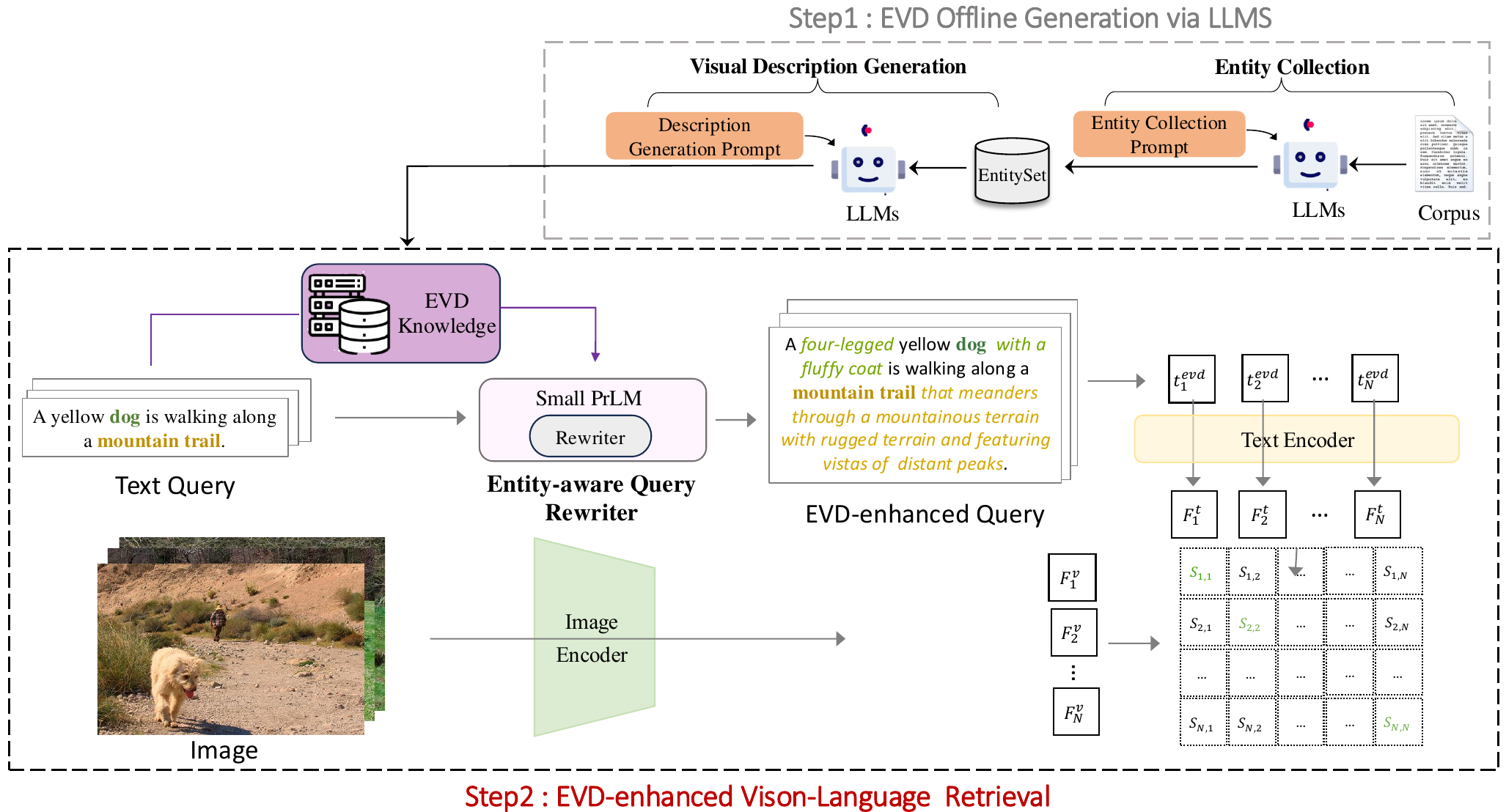}
  \caption{The overall architecture of EvdCLIP comprises two components: EVD offline generation via LLMs and EVD-enhanced vision-language retrieval. First, EVD knowledge is generated offline using LLMs. Then, an EVD-aware query rewriter integrates the query with EVD to produce an EVD-enhanced query for retrieval.
  }
  \label{fig:architecture}
\end{figure*}

\section{Related Work}
\subsection{Vision-Language Retrieval}
Previous VLR models fall into three categories: single-stream, double-stream, and dual-encoder. \textbf{Single-stream} models~\cite{chen2020uniter, kim2021vilt} use self-attention for fine-grained multi-modal alignment. \textbf{Double-stream} models~\cite{li2021align, li2022blip, 10.1145/3581783.3612507, yang2022vision, zeng2022multi} process intra-modality features with a shared fusion encoder, decoupling intra-modal and cross-modal modeling. Due to the need for efficient inference in visual language retrieval, \textbf{dual-encoder} architectures~\cite{radford2021learning, xie2022token} have been proposed, using contrastive learning to align visual and text embeddings in the same semantic space. To enhance image-text alignment, we introduce the EvdCLIP framework, which integrates entity visual descriptions as alignment cues.

\subsection{Knowledge Acquisition for VLR}
Multi-modal data often includes implicit information, making knowledge extraction essential. Related work falls into two categories: internal knowledge mining and external knowledge incorporation.
\textbf{Internal Knowledge Mining}: 
Methods like object detection and statistics mine knowledge from the data. For example, OA-Trans~\cite{wang2022object} and structureCLIP~\cite{huang2024structure} use objects in images for cross-modal learning, while Coder~\cite{wang2022coder} and ViSTA~\cite{cheng2022vista} leverage common knowledge and scene text for image-text retrieval.
\textbf{External Knowledge Incorporation}:
Some methods integrate external knowledge to enhance VLR models. Knowledge-CLIP~\cite{pan2022contrastive} and ACP~\cite{pan2022contrastive} use multi-modal knowledge graphs to improve concept-level semantics. EI-CLIP~\cite{ma2022ei} extends entity semantics through e-commerce knowledge for better e-commerce retrieval.
\par
LLMs can be considered as vast knowledge bases~\cite{zeng2022glm,menon2022visual}. In this work, we explore the use of the rich knowledge in LLMs to enhance image-text alignment, improving MMR performance.

\subsection{Description-enhanced for CLIP}
Recent work has focused on enhancing CLIP using category descriptions in image classification and object detection. 
For instance, \cite{menon2022visual} and \cite{pratt2023does} generate descriptions with LLMs, while \cite{yao2022detclip} improves object detection through parallel training with an object concept dictionary. As noise in description has gained attention, filtering methods have emerged. \cite{yang2023language} designs a scoring function to select representative descriptions and uses a learnable weight matrix for personalized attention. \cite{an2023more} uses a manually designed scoring function to reflect annotators' linguistic preferences, focusing on relevant features. \cite{maniparambil2023enhancing} addresses interfering information with a self-attention adapter.
\par

However, these works focus on image classification and detection, while EvdCLIP is tailored for VLR. In VLR, integration of complex queries and EVDs faces two challenges: dynamic filtering of EVD noise based on query and high-quality EVD-enhanced query generation. To address this, we developed EaRw, a rewriter that dynamically selects relevant descriptions, generating fluent, high-quality queries.

\section{Methodology}
The framework of EvdCLIP is illustrated in Figure \ref{fig:architecture}. We first review the dual-encoder framework, and then detail EVD offline generation via LLMs.  Finally, we illustrate how we utilize EVD  to enhance multimodal retrieval.
\par
\subsection{Dual-encoder Framework}
In this work, we select the simple yet effective dual-encoder CLIP as our backbone. As shown in Figure \ref{fig:architecture}, images and texts are encoded by an image encoder and a text encoder respectively, then projected into the same semantic space for effective retrieval.
Formally, assuming we have $N$ samples in a batch, $B = \left \{ \left ( v_i, t_i \right )\right \}_{i=1}^{N}$ denotes the training dataset, where $\left ( v_{i},t_{i}\right )$ is the $i$-th image-text pair. The matched image-text pairs are considered positive samples, while other pairwise combinations serve as negative samples. We define the image-to-text contrastive loss as:
% \small{
\begin{equation}
\begin{split}
L_{i2t} &= -\frac{1}{N}\sum_{\left ( v_{i},t_{i}\right )\in B}y\cdot \text{log} p\left ( v_{i},t_{i}\right ) \\ 
 &=  -\frac{1}{N}\sum_{\left ( v_{i},t_{i}\right )\in B}\text{log}\frac{\text{exp}\left ( F_{i}^{v}\cdot F_{i}^{t} /\tau  \right )}{\sum_{j=0}^{N}\text{exp}\left ( F_{i}^{v}\cdot F_{j}^{t}/\tau  \right )}.
\end{split}
\end{equation}
% }
where $F_{i}^{v}$ and $F_{i}^{t}$ are the normalized embedding of $v_i$ and $t_i$. $\tau$ is the temperature hyper-parameter. Similarly, we can define the text-to-image contrastive loss as:

\begin{equation}
L_{t2i}= -\frac{1}{N}\sum_{\left ( v_{i},t_{i}\right )\in B}\text{log}\frac{\text{exp}\left ( F_{i}^{v}\cdot F_{i}^{t} /\tau  \right )}{\sum_{j=0}^{N}\text{exp}\left ( F_{j}^{v}\cdot F_{i}^{t}/\tau  \right )}.
\end{equation}

The final contrastive loss can be denoted as:
\begin{equation}
L = L_{i2t} + L_{t2i}
\label{eq:loss}
\end{equation}
\par
The dual-encoder framework aligns images and text using global features, but it lacks fine-grained cues for precise vision-language alignment. To address this, we use EVD as additional cues, enhancing retrieval performance.

\subsection{EVD Offline Generation via LLMs}

\subsubsection{\textbf{Entity Collection}} 
To build the EVD knowledge base, we first create a predefined entity set. In VLR, visual items with rich visual information are more critical than non-visual terms. For example, non-visual terms like "New York" contribute little to image-text retrieval due to the difficulty in concisely describing its visual characteristics. In contrast, visual terms like "whale" and "school bus" offer distinct visual cues that enhance cross-modal retrieval.
\par
We collect visual items from the training datasets of Flickr30k~\cite{plummer2015flickr30k} and MSCOCO~\cite{lin2014microsoft}. Since current methods struggle to differentiate between visual and non-visual entities, we use LLMs with carefully designed prompts to extract visual entities.
Specifically, the prompts clarify the distinction between visual and non-visual entities, as detailed in Appendix~\ref{appendix_prompt}, allowing LLMs to accurately extract visual entities from the text. To ensure precise and standardized extraction, we include two QA examples in the prompt.
The final entity set can be denoted as $E=\left \{{e_{n}} \right \}_{n=1}^{M}$, where $M$ indicates the number of entities and $e_n$ represents the $n$-the entity name.

\subsubsection{\textbf{Visual Description Generation}}
Given the entity set, we use a large language model to generate visual descriptions focused on distinguishable features like shape and color, facilitating fine-grained cross-modal alignment. We design an instruction prompt, detailed in Appendix~\ref{appendix_prompt}, where {entity names} represent entities from set $E$.

% This approach yields a list of visual descriptions for each entity, as shown in Figure~\ref{fig:evd_case}, emphasizing characteristics such as color, shape, parts, and quantity to enhance visual distinction. Leveraging the knowledge in large language models, these descriptions are diverse and detailed.
% The EVD knowledge base $O=\left \{e_i:{evd}_i \right \}_{i=1}^{M}$ maps each entity $e_i$ to its corresponding visual descriptions ${evd}_i$, covering around 10,000 entities in this paper. Here ${evd}_i$ represents a list of multiple visual descriptions of entity $e_i$

This approach generates a list of visual descriptions for each entity, focusing on characteristics like color, shape, parts, and quantity to enhance visual distinction (examples in Appendix~\ref{app:exaple_evd}). For entities with multiple meanings, we discuss description generation in detail in Appendix~\ref{app:multi-mean}.
The EVD knowledge base $O=\left \{e_i:{evd}_i \right \}_{i=1}^{M}$ maps each entity $e_i$ to its corresponding visual descriptions ${evd}_i$, covering around 10,000 entities in this paper. Here ${evd}_i$ represents a list of multiple visual descriptions of entity $e_i$.
Once the EVD knowledge base is constructed offline, there is no need to generate EVD during either training or inference.

% \begin{figure}[t]
% \centering
% \includegraphics[width=\linewidth]{Figures/evd_case.pdf}
% \caption{Examples of entity visual descriptions.}
% \label{fig:evd_case}
% \end{figure}

\subsection{EVD-aware Rewriter}

Given the query $t_i$, we first retrieve the entities $e_i$ and obtain their descriptions ${evd}_i$ from the EVD knowledge base. The EVD-enhanced query is then formed as $t_i^{evd} = agg(t_i, {evd}_i)$, where $agg(\cdot )$ represents the integration strategy. As shown in Figure~\ref{fig:noise_sample}, existing methods face noise and low-quality issues. To overcome these challenges, we develop the EVD-aware Rewriter (EaRW), which uses a pre-trained language model to expand queries with EVD knowledge. As shown in Figure~\ref{fig:architecture}, before multimodal retrieval, the T5 model rewrites the query by integrating it with its corresponding EVD, generating the EVD-enhanced query:

% To ensure the T5 model can effectively enhances query with EVD knowledge, we design a rewriter's prompt template. This template provides clear instructions to the rewriter.

\subsubsection{\textbf{EVD-enhanced Query Rewriting Dataset}}

However, the multimodal  knowledge-enhanced rewriting introduces gaps with T5's pre-training, causing EaRW to sometimes struggle with the rewriting task, limiting its performance. To better filter EVD noise and improve integration quality, we propose a specialized training scheme for the T5 model. First, we construct an EVD-enhanced query rewriting dataset $D_{EQR}$. 
Inspired by recent distillation methods~\cite{ma2023query}, we use LLMs to rewrite queries and collect EVD-enhanced queries with positive feedback from CLIP as pseudo-labels in the training dataset $D_{EQR}$. For details and instructions on constructing the dataset $D_{EQR}$ can be seen in Appendix~\ref{warmup_instruct}. 
We generate multiple EVD-enhanced queries for each query, and the final $D_{EQR}$ is composed of tuples $(x: \{y_i, s_i\}_{i=1}^k)$, where $x$ is the original query, $y_i$ is the $i$-th EVD-enhanced query label for $x$, $s_i$ is the corresponding score, and $k$ is the number of pseudo-labels for each $x$.
In summary, we leverage ChatGPT's contextual reasoning ability and CLIP's  feedback to generate a high-quality corpus that effectively captures Context preferences and dataset preferences~\cite{dunlap2024describing} of entities.

% For the detail, Our dataset is constructed as shown in Figure~\ref{fig:warmupdata_case}. LLMs serve as a teacher for the EaRW.  We generate multiple ($k$) candidate EVD-enhanced queries for each original query by using prompt templates in the Figure~\ref{fig:prompt}. These candidate sets vary in quality. Considering that high-quality and low-noise EVD-enhanced queries enhance the CLIP model's retrieval accuracy, we input these along with the original query into CLIP and calculate the matched score with the corresponding images. Samples that score higher than the original query are selected for the evd-enhanced query rewriting dataset $D_{Train}$. Specifically speaking, We employ GPT-3.5 as a teacher model to generate the candidate EVD-enhanced queries for each original query. Subsequently, we use a pre-trained CLIP model to evaluate and select the top-5 effective EVD-enhanced query and their matched score. In this way, We form a tuple $(x, y_w, y_l)$ of each effective EVD-enhanced query and its corresponding score, where $x$ is the original query, $y_w$ is the corresponding rewritten text, and $y_l$ is the score.
% We use the training set of benchmark dataset (e.g., Flickr30k or MSCOCO) to create the warm-up dataset's training set, and similarly, we use the test set to generate the warm-up dataset's test set. In summary, we utilize ChatGPT's powerful contextual reasoning capabilities and CLIP's scoring feedback to generate a high-quality corpus, which is effective in capturing entities' sample preferences and dataset preferences. 

\subsubsection{\textbf{Rewriter Warm-up}}	
We initiate the EaRW with a pre-trained T5-large model. The rewriter is first trained on rewriting dataset $D_{EQR}$ to warm up. In this step, we use the dataset $D_{EQR}$ to train an initial rewriter via a supervised fine-tuning method. This process as a text-to-text task and the rewriter is finetuned on $D_{EQR}$ with the standard log-likelihood as the training objective, denoted as:

% \begin{equation}
%     L_{\text{SFT}} = - \sum_{t} \log p_{\theta}(\hat{\tilde{x}}_t | \tilde{x}_{<t}, x)
% \end{equation}
\begin{equation}
    L_{\text{SFT}}(\theta) = -\mathbb{E}_{(x, y) \sim D_{EQR}} \sum_{t} \log \pi \left(\hat{y}_t |  \hat{y}_{<t}, x; \theta \right)
\end{equation}

where $x$ refers to the original query and $\tilde{y}$ refers to the corresponding EVD-enhanced query label with the highest score.  $\pi(\cdot )$ and $\theta$ denote our query rewriter and its parameters. The performance of EaRw after warm-up may be sub-optimal. In order to better align the EaRw with the retriever CLIP, we further employ preference optimization~\cite{peng2024large} to fine-tune the EaRw to fit the retriever.

\begin{table*}

  \caption{Fine-tuning results for image-text retrieval on the Flickr30K (1K) test set and MSCOCO (5K) test set. 
  Notations: V-Encoder: vision encoder; \# PT Data: the pre-training datasets.}
  \label{ablation-overall}
  \centering
 \resizebox{0.9\linewidth}{!}{
  \begin{tabular}{llllll|lll|lll|lll}
     \hline
     \multirow{3}{*}{Methods}  & \multirow{3}{*}{V-Encoder} &  \multirow{3}{*}{\# PT Data} &   \multicolumn{6}{c}{Flickr30K(1K)} & \multicolumn{6}{c}{MSCOCO(5K)}  \\
     &  & & \multicolumn{3}{c}{I2T Retrieval} & \multicolumn{3}{c}{T2I Retrieval} & \multicolumn{3}{c}{I2T Retrieval} & \multicolumn{3}{c}{T2I Retrieval} \\
     &  & & \multicolumn{1}{|c}{R@1} & R@5 & R@10 & R@1 & R@5 & R@10 & R@1 & R@5 & R@10 & R@1 & R@5 & R@10 \\
     \hline
     CLIP~\cite{radford2021learning}   & ViT-B/32 & NA &  \multicolumn{1}{|c}{64.8} &  85.7 & 92.5  & 49.2 & 79.3 & 86.8 & 43.7 & 73.5 & 82.6 & 32.7 & 63.3 & 75.0\\
     DetCLIP~\cite{yao2022detclip}   & ViT-B/32 & NA &  \multicolumn{1}{|c}{65.2} &  86.3 & 93.5  & 50.7 & 79.2 & 86.8 & 45.2 & 73.7 & 83.4 & 33.4 & 63.5 & 75.0\\     
     DesCLIP~\cite{menon2022visual}   & ViT-B/32 & NA &  \multicolumn{1}{|c}{65.8} &  87.7&93.6&51.2&79.8&87.1&45.7&73.9&83.8&34.2&63.8&75.2\\
     CLIP-GPT~\cite{maniparambil2023enhancing}   & ViT-B/32 & NA &  \multicolumn{1}{|c}{66.5} & 88.1&93.6&51.2&80.1&\textbf{87.8}&46.1&74.0&83.7&34.1&63.7&75.3\\
     LaBo~\cite{yang2023language}   & ViT-B/32 & NA &  \multicolumn{1}{|c}{66.1} &  87.5&93.5&51.2&79.8&87.5&46.4&74.1&83.8&34.3&63.7&75.1\\
     EvdCLIP & ViT-B/32 & NA & \multicolumn{1}{|c}{\textbf{66.9}} & \textbf{88.6} & \textbf{94.2} & \textbf{52.0} & \textbf{80.5} & 87.6 & \textbf{46.8} & \textbf{74.4} & \textbf{84.2} & \textbf{35.2} & \textbf{64.5} & \textbf{75.7}\\

     % CLIP (Zero-shot) & Laion400M & 80.8 & 94.7 & 97.8 & 62.6 & 85.6 & 91.2 & 53.2 & 77.4 & 85.5  & 36.2 & 61.6 & 71.8\\
     \hline
     CLIP~\cite{radford2021learning}  & ViT-B/32 & Laion400M & \multicolumn{1}{|c}{89.1} & 97.8 & 98.9 & 74.1 & 92.6 & 95.9 & 65.3 & 85.9 & 91.9 & 48.1 & 75.0  & 83.7\\
     DetCLIP~\cite{yao2022detclip}   & ViT-B/32 & Laion400M &  \multicolumn{1}{|c}{89.2} &  97.8&99.1&74.6&92.8&96.0&65.5&85.9&92.1&48.3&75.1&83.7\\     
     DesCLIP~\cite{menon2022visual}   & ViT-B/32 & Laion400M &  \multicolumn{1}{|c}{89.6} &  98.6&99.3&75.1&93.0&95.9&66.1&86.1&92.4&48.8&75.3&84.1\\
     CLIP-GPT~\cite{maniparambil2023enhancing}   & ViT-B/32 & Laion400M &  \multicolumn{1}{|c}{89.7} &  98.7&99.2&75.2&93.1&96.1&66.2&86.2&92.3&48.8&75.3&84.3\\
     LaBo~\cite{yang2023language}   & ViT-B/32 & Laion400M &  \multicolumn{1}{|c}{89.7} &  98.5&99.2&74.8&93.1&96.0&66.3&86.1&\textbf{92.6}&49.0&75.2&84.2\\   
     EvdCLIP & ViT-B/32 & Laion400M & \multicolumn{1}{|c}{\textbf{90.7}} & \textbf{99.1} & \textbf{99.5} & \textbf{75.6} & \textbf{93.5} & \textbf{96.5} & \textbf{66.8} & \textbf{86.8} & 92.6 & \textbf{49.5} & \textbf{75.8} & \textbf{84.5} \\

     \hline
     
     CoCa~\cite{yu2022coca} & ViT-B/32 & Laion-2B & \multicolumn{1}{|c}{85.5} & 96.5 & 98.7 & 72.0 & 91.2 & 95.4 & 63.9 & 85.6 & 91.0 & 45.6 & 72.1  & 82.2 \\
     DetCoCa~\cite{yao2022detclip}   & ViT-B/32 & Laion-2B &  \multicolumn{1}{|c}{85.6} &  96.5&98.7&72.2&91.2&95.4&63.8&85.5&91.0&45.8&72.1&82.1\\     
     DesCoCa~\cite{menon2022visual}   & ViT-B/32 & Laion-2B &  \multicolumn{1}{|c}{86.2} &  96.8&98.9&72.3&91.4&95.4&64.2&\textbf{85.7}&91.2&46.0&72.3&82.2\\
     CoCa-GPT~\cite{maniparambil2023enhancing}   & ViT-B/32 & Laion-2B &  \multicolumn{1}{|c}{86.2} & 97.0&98.8&72.2&\textbf{91.6}&95.3&64.3&85.7&91.0&46.0&72.2&82.3\\
     LaBo~\cite{yang2023language}   & ViT-B/32 & Laion-2B &  \multicolumn{1}{|c}{86.1} &  96.8&98.8&72.1&91.5&95.5&64.3&85.6&91.1&46.1&72.1&82.3\\ 
     EvdCoCa & ViT-B/32 & Laion-2B & \multicolumn{1}{|c}{\textbf{86.6}} & \textbf{97.2} & \textbf{98.9 }& \textbf{72.6} & 91.5 & \textbf{95.7} & \textbf{64.8} & 85.7 & \textbf{91.5} & \textbf{46.4} & \textbf{72.6}  & \textbf{82.5} \\
     
     % \hline
     % BLIP-2~\cite{li2023blip} & ViT-L/14 & Laion400M & \multicolumn{1}{|c}{} \\
     % DetBLIP-2~\cite{yao2022detclip}   & ViT-L/14 & Laion400M & \multicolumn{1}{|c}{}  \\ 
     % DesBLIP-2~\cite{menon2022visual}   & ViT-L/14 & Laion400M & \multicolumn{1}{|c}{}  \\
     % BLIP-2-GPT~\cite{maniparambil2023enhancing}   & ViT-L/14 & Laion400M & \multicolumn{1}{|c}{} \\
     % LaBo~\cite{yang2023language}   & ViT-L/14 & Laion400M & \multicolumn{1}{|c}{} \\   
     % EvdBLIP-2~ & ViT-L/14 & Laion400M & \multicolumn{1}{|c}{} \\

     \hline
     EVA-02-CLIP~\cite{sun2023eva} & ViT-B/16 & Merged-2B & \multicolumn{1}{|c}{90.8} & 98.7&99.2&78.9&94.7&97.0&69.1&89.2&94.0&52.6&78.5&86.8\\
     DetEVA-02-CLIP~\cite{yao2022detclip}   & ViT-B/16 & Merged-2B &  \multicolumn{1}{|c}{90.9} &  98.6&99.1&79.1&94.6&97.0&69.3&89.2&94.0&52.7&78.5&86.7\\     
     DesEVA-02-CLIP~\cite{menon2022visual}   & ViT-B/16 & Merged-2B &  \multicolumn{1}{|c}{91.1} &  \textbf{98.7}&99.2&79.3&94.7&97.1&69.5&89.3&94.3&52.6&78.6&86.8\\
     EVA-02-CLIP-GPT~\cite{maniparambil2023enhancing}   & ViT-B/16 & Merged-2B &  \multicolumn{1}{|c}{91.1} &  98.7&99.2&79.3&94.7&97.1&69.4&89.3&94.3&52.6&78.6&86.8\\
     LaBo~\cite{yang2023language}   & ViT-B/16 & Merged-2B &  \multicolumn{1}{|c}{91.0} &  98.6&99.3&79.3&94.8&97.0&69.4&89.2&94.1&52.8&78.5&86.8\\
     EvdEVA-02-CLIP & ViT-B/16 & Merged-2B & \multicolumn{1}{|c}{\textbf{91.4}} & 98.6 & \textbf{99.5} & \textbf{79.7} & \textbf{94.8} & \textbf{97.2}& \textbf{69.9} & \textbf{89.7} & \textbf{94.5} & \textbf{53.4} & \textbf{78.9} & \textbf{87.1} \\

     \hline
  \end{tabular}
  }
\end{table*}

\begin{table}
  \centering
  \caption{Fine-tuning T2I retrieval results on HuaWei Business Datasets. The vision encoder is ViT-B/32.}
  \label{tab:huawei}
  \centering
  % \begin{subtable}[c]{1.\linewidth}
    \centering
    \resizebox{0.8\linewidth}{!}{
    \begin{tabular}{c|ccc|ccc}
     \hline
     \multirow{2}{*}{Methods} & \multicolumn{3}{c|}{Theme} & \multicolumn{3}{c}{Wallpaper}  \\
     & R@5 & R@50 & R@100 & R@5 & R@50 & R@100  \\
     \hline
     CLIP & 50.32 & 64.22  & 67.68 & 22.30 & 52.01 & 62.41   \\
     EvdCLIP & \textbf{50.47} & \textbf{67.30 }& \textbf{71.85} & \textbf{25.22} & \textbf{58.71} & \textbf{69.13 }\\
     $\triangle$ & +0.15 & +3.08 & +4.17 & +2.92 & +6.70 & +6.72  \\
     \hline 
    \end{tabular}}
    % \caption{Subtable A}
  % \end{subtable}

  \vspace{.1cm} % Adjust vertical space between subtables

  % \begin{subtable}[c]{1.\linewidth}
    \centering
    \resizebox{0.8\linewidth}{!}{
    \begin{tabular}{c|ccc|ccc}
     \hline
     \multirow{2}{*}{Methods} & \multicolumn{3}{c|}{Lock-Screen} & \multicolumn{3}{c}{Icons}  \\
     & R@5 & R@50 & R@100 & R@5 & R@50 & R@100  \\
     \hline
     CLIP & 83.51 & 92.46 & 94.31 &  73.97 & 86.84 & 89.71   \\
     EvdCLIP & \textbf{84.73} & \textbf{94.50} & \textbf{95.93} & \textbf{74.03} & \textbf{87.38} & \textbf{90.41}\\
     $\triangle$ & +1.22 & +2.04 & +1.62 & +0.06 & +0.54 & +0.70  \\
     \hline
    \end{tabular}}
    % \caption{Subtable B}
  % \end{subtable}
\end{table}

\subsubsection{\textbf{Preference Alignment}}

% The rewriter model after warm-up shows modest performance. To further enhance the EVD-enhanced multimodal retrieval framework, we propose joint training of the rewriter and CLIP. Specifically, we will set the EaRW before the text encoder and rewrite the original query. To ensure that CLIP can understand queries with varying descriptive granularities, we randomly perform the query rewrite during the training process.
% Subsequently, we will jointly optimize the CLIP and the rewriter in the whole framework according to (Eq.~\ref{eq:loss}).
This process requires the construction of a specialized preference dataset. As detailed in Appendix~\ref{warmup_instruct}, we generate multiple EVD-enhanced queries and obtain image-text similarity scores from the retrieval system, which serve as rewards for preference learning. These scores allow us to rank the EVD-enhanced queries from highest to lowest preference. To minimize bias from the reward model and enhance fine-grained preference comparisons from a global perspective, we introduce Preference Rank Optimization (PRO) based on the Bradley-Terry model\cite{song2024preference}. This method guides the model to learn the ranking of rewrites according to feedback from the retriever. According to the Bradley-Terry model, the probability of choosing a policy is proportional to its corresponding reward. Given the partial order relation $y_1 \succ y_2$, the preference probability can be expressed as:
\begin{equation}
    P_{BT} = \frac{\exp(r(y_1, x))}{\exp(r(y_1, x)) + \exp(r(y_2, x))}
\end{equation}
where $r(\cdot)$ is the reward function, which is defined as the normalized log probability of the rewrite generated in PRO. PRO extends pairwise partial order into general listwise partial order. The PRO loss is expressed by the equation:

\begin{equation}
    L_{\text{PRO}}(\theta) = -\mathbb{E}_{(\mathbf{x}, \mathbf{y}) \sim D_{EQR}} 
\sum_{j=1}^{k-1} \log \frac{\exp\left(\frac{\pi_{\text{PRO}}(y_j | x; \theta)}{T_{j}^{j}}\right)}
{\sum_{i=j}^{k} \exp\left(\frac{\pi_{\text{PRO}}(y_i | x; \theta)}{T_{j}^{i}}\right)}
\end{equation}
where $T_{j}^{i} = \frac{1}{r(y_j) - r(y_i)}$ and $T_{j}^{j} = \min_{i > j} (T_{j}^{i})$ are used to measure ranking difference. $k$ denotes the number of candidate Evd-enhanced query label, $\pi_{\text{PRO}}$ and $\theta$ refer to the policy model and its parameters. Additionally, an SFT loss is applied to the PRO loss with weight $\beta$ to preserve the model's ability to generate standard outputs.
\begin{equation}
    L_{\text{ALIGN}} = L_{\text{PRO}} + \beta L_{\text{SFT}}
\end{equation}

EaRW not only learns to recognize and integrate relevant visual descriptions based on entity preferences but also harnesses LLM's ability to generate fluent, high-quality queries. This method effectively mitigates the issues of noise and low-quality of EVD-enhanced query.

We integrate the optimized EaRW into the CLIP framework, fine-tuning CLIP using Eq.~(\ref{eq:loss}) while keeping EaRW's parameters frozen. To handle queries with varying descriptive granularities, we randomly apply query rewriting with probability $p$ during training. For inference, we average the EVD-enhanced query score with the original query score to determine the final score.

\section{Experiments}

\subsection{Experimental Setup}
\subsubsection{\textbf{Datasets}.} 
This paper utilizes four types of datasets: pre-training datasets, benchmark datasets, Huawei business datasets, and EVD-enhanced query rewriting dataset $D_{EQR}$. We use the benchmark and Huawei business datasets for model fine-tuning and performance evaluation.
\par
\textbf{Pre-training Datasets}: (1) Laion400M~\cite{schuhmann2021laion} and (2) Laion-2B~\cite{schuhmann2021laion} contain 400 million and 2 billion image-text pairs respectively, sourced from publicly available internet data. (3) Merged-2B~\cite{sun2023eva} combines multiple datasets, totaling 2 billion image-text pairs. (4) YFCC15M~\cite{thomee2016yfcc100m} is a subset of YFCC100M, with 15 million image-text pairs. Finally, (5) CC12M~\cite{changpinyo2021conceptual} consists of 12 million image-text pairs designed for pre-training.

\par
\textbf{Benchmark Datasets}: 
(1) Flickr30K~\cite{plummer2015flickr30k}  contains 31,000 images, each annotated with 5 captions. Following~\cite{li2021align}, which split into 29K/1k/1k images for training, validation and testing. 
(2) MSCOCO~\cite{lin2014microsoft} comprises 123,287 images,  each annotated with 5 captions. We split it into 114K/5K/5K for training, validation, and testing. 
(3) MSR-VTT~\cite{xu2016msr} includes 10K videos, each with 200K text. We employ 9K videos for training and evaluation on the 1K test set.
(4) SBU30k~\cite{ordonez2011im2text} consists of 36k image-text pairs, randomly sampled from SBU Captions and split into 30K/3K/3K for training, validation, and testing. Similarly, we obtain (6) CC30K and (7) YFCC30K by randomly sampling from CC12M and YFCC15M.

\par
\textbf{Huawei Business Datasets}: This dataset, sourced from Huawei Mobile Scene Search Service, contains a large number of Chinese image-text pairs. It is categorized into four types: Theme, Wallpaper, Lock-Screen, and Icon. Detailed information and statistics are provided in Appendix~\ref{huawei data}.
% (1) Thema dataset comprises image-text pairs related to mobile phone themes.
% (2) Wallpaper dataset consists of image-text pairs representing mobile phone wallpapers.
% (3) Lock-Screen dataset incorporates image-text pairs designed for mobile phone lock screens.
% (4) Icon dataset is collected from image-text pairs representing mobile phone icons.
% Table~\ref{ablation-dataset} provides their statistics, including sample quantities for the train/validation/test, 
% as well as the average token length and the average number of entities in text queries.
\par
% \textbf{Warm-up Datasets}: 
% The warm-up dataset construction approach and details can be found in the Appendix~\ref{warmup_instruct}
\par

\subsubsection{\textbf{Baseline}}
We will validate our approach on advanced dual-encoder retrieval models:
(1) CLIP~\cite{radford2021learning}, a powerful dual-encoder model pre-trained with contrast learning. (2) CoCa~\cite{yu2022coca}, a framework that integrates various pre-training paradigms, using its image encoder and unimodal text decoder for retrieval. (3) EVA-02-CLIP~\cite{sun2023eva}, which incorporates novel techniques for representation learning,  enhancing CLIP's performance.

\par
We also compare EvdCLIP with description-enhanced CLIP methods:
(1) DetCLIP\cite{yao2022detclip} generates object concepts via WordNet.
(2) DesCLIP~\cite{menon2022visual} uses LLMs to generate descriptions and inputs them into CLIP in parallel.
(3) CLIP-GPT~\cite{maniparambil2023enhancing} creates visual descriptions with LLMs and denoising with a self-attention adapter.
(4) LaBo~\cite{yang2023language} selects descriptions with designed functions and a learnable weighting matrix.
All methods are our implementations in VLR, details are shown in Appendix~\ref{appendix_detail}.

\subsubsection{\textbf{Large Language Models}}
We used several LLMs in our research, including GPT-3~\cite{brown2020language} ("text-davinci-003"), ChatGPT~\cite{openai2022chatgpt} ("GPT-3.5-turbo"), Llama~\cite{touvron2023llama} ("Llama-2-13B-chat"), Vicuna~\cite{chiang2023vicuna} ("vicuna-13B-v1.5"), and PanGu~\cite{zeng2104pangu}, a Chinese LLM developed by Huawei.

\subsubsection{\textbf{Implementation Details}} 
EvdCLIP is fine-tuned directly on the pre-trained CLIP without re-pretraining, making the process lightweight. Details can be found in Appendix~\ref{appendix_detail}.

\subsection{Main Results}
% We first evaluate the effectiveness of our approach on the state-of-the-art dual-encoder framework for image-text retrieval, using two widely-used benchmark datasets: Flickr30K and MSCOCO. As demonstrated in Table~\ref{ablation-overall}, EvdCLIP consistently outperforms the CLIP model across all metrics on both Flickr30K and MSCOCO. It illustrates that incorporating EVD as additional cues enhances the model's ability to semantically align images and text. Notably, EvdCLIP shows a significant boost in the R@$1$ metric. When pre-trained on Laion400M, EvdCLIP achieves an R@$1$ increase of 1.5\%, 1.4\%, 1.3\%, and 1.2\% on I2T and T2I for Flickr30K and MSCOCO, respectively. This suggests that EVD effectively captures fine-grained differences among entities, enabling more precise identification of entities.
We evaluate our approach on a state-of-the-art dual-encoder framework for VLR using two benchmarks: Flickr30K and MSCOCO. As shown in Table~\ref{ablation-overall}, EvdCLIP consistently outperforms CLIP across all metrics on both datasets, demonstrating that incorporating EVD enhances the alignment of images and text. Notably, EvdCLIP shows a significant improvement on R@$1$. When pre-trained on Laion400M, EvdCLIP achieves R@$1$ increases of 1.6\%, 1.5\%, 1.5\%, and 1.4\% on I2T and T2I for Flickr30K and MSCOCO, respectively, indicating that EVD captures fine-grained entity differences, leading to more precise identification.
\par
% From Table \ref{ablation-overall}, we observe that our approach improves more significantly on the R$@1$ metric. 

% thus can better identify the indistinguishable entities.
\par
% We also applied our method to other CLIP-style models. As shown in Table~\ref{ablation-overall}, both CoCa and EVA-02-CLIP, when enhanced with our approach, achieved superior performance across most metrics, confirming its compatibility and effectiveness across various dual-encoder frameworks. Even though EVA-02-CLIP significantly enhances CLIP's performance through optimization strategies, our approach further improves its effectiveness. Compared to existing description enhancement-based methods, our method is tailored for VLR and thus improves retrieval performance more significantly. Detailed analysis can be found in Ablation studies.
We test EvdCLIP on other CLIP-style models. As shown in Table~\ref{ablation-overall}, both CoCa and EVA-02-CLIP, with our approach, achieve superior performance across most metrics, demonstrating its compatibility and effectiveness. Although EVA-02-CLIP already significantly improves CLIP's performance through optimization strategies, our method further enhances its performance. Compared to existing description enhancement methods, EvdCLIP is tailored for VLR, leading to more significant improvements in retrieval performance. Detailed analysis is provided in ablation studies.
\par
% Furthermore, we find that EvdCLIP* achieves comparable performance to EvdCLIP. This indicates that the EvdCLIP model, trained with EVD, possesses a better understanding of the visual semantics of entities. Therefore, the model can still perform entity-image alignment even without EVD knowledge during inference.
%没有说后面都不带，那就是后面都带了，这个得说清楚
\subsection{Results on Huawei Business Dataset}
We also evaluate our method on Huawei business dataset and the results are consistent with those from public datasets. Using the Pangu Chinese LLM to generate entity visual descriptions, EvdCLIP consistently outperforms CLIP in various text-to-image retrieval tasks, as shown in Table~\ref{tab:huawei}. Notably, we observe that our method achieves the most significant performance gains in the wallpaper task, with recall rates improving by 2.92\%, 6.70\%, and 6.72\% at R@5, R@50, and R@100. We speculate that user queries for the wallpaper are often short, vague, and entity-rich, making EVDs particularly crucial for this task. These results further demonstrate the effectiveness of our EVDs in large-scale Chinese vision-language retrieval.
\par

\begin{table}
  \caption{Ablation studies on description sources.
  The vision encoder is ViT-B/32, Fine-tuning dataset is Flickr30k and Pre-Training dataset is Laion400M.}
  \label{ablation-4}
  \centering
\adjustbox{width=1\linewidth}{
  \begin{tabular}{l|l|lll|lll}
     \hline
     \multirow{2}{*}{Methods} & \multirow{2}{*}{Des. Source} & \multicolumn{3}{c|}{I2T Retrieval} & \multicolumn{3}{c}{T2I Retrieval}  \\
     & & R@1 & R@5 & R@10 & R@1 & R@5 & R@10 \\
     \hline
     CLIP & NA & 89.1 & 97.8 & 98.9 & 74.1 & 92.6 & 95.9 \\
     \hline
     WordNetCLIP & WordNet & 89.2 & 97.8 & 99.2 & 74.6 & 92.8 & 96.0 \\
     \hline
     \multirow{4}{*}{EvdCLIP} & GPT-3 & 90.6 & 99.0 & 99.4 & \textbf{75.6} & 93.4 & 96.4 \\
     & ChatGPT & \textbf{90.7} & 99.1 & \textbf{99.5} & 75.6 & \textbf{93.5} & \textbf{96.5} \\ 
     & Llama-13B & 90.4 & 98.8 & 99.4 & 75.3 & 93.3 & 96.2 \\ 
     & Vicuna-13B & 90.2 & 98.8 & 99.2 & 75.4 & 93.2 & 96.3 \\      
     \hline
  \end{tabular}}
\end{table}

\begin{figure}
  \centering
  \includegraphics[width=1.0\linewidth]{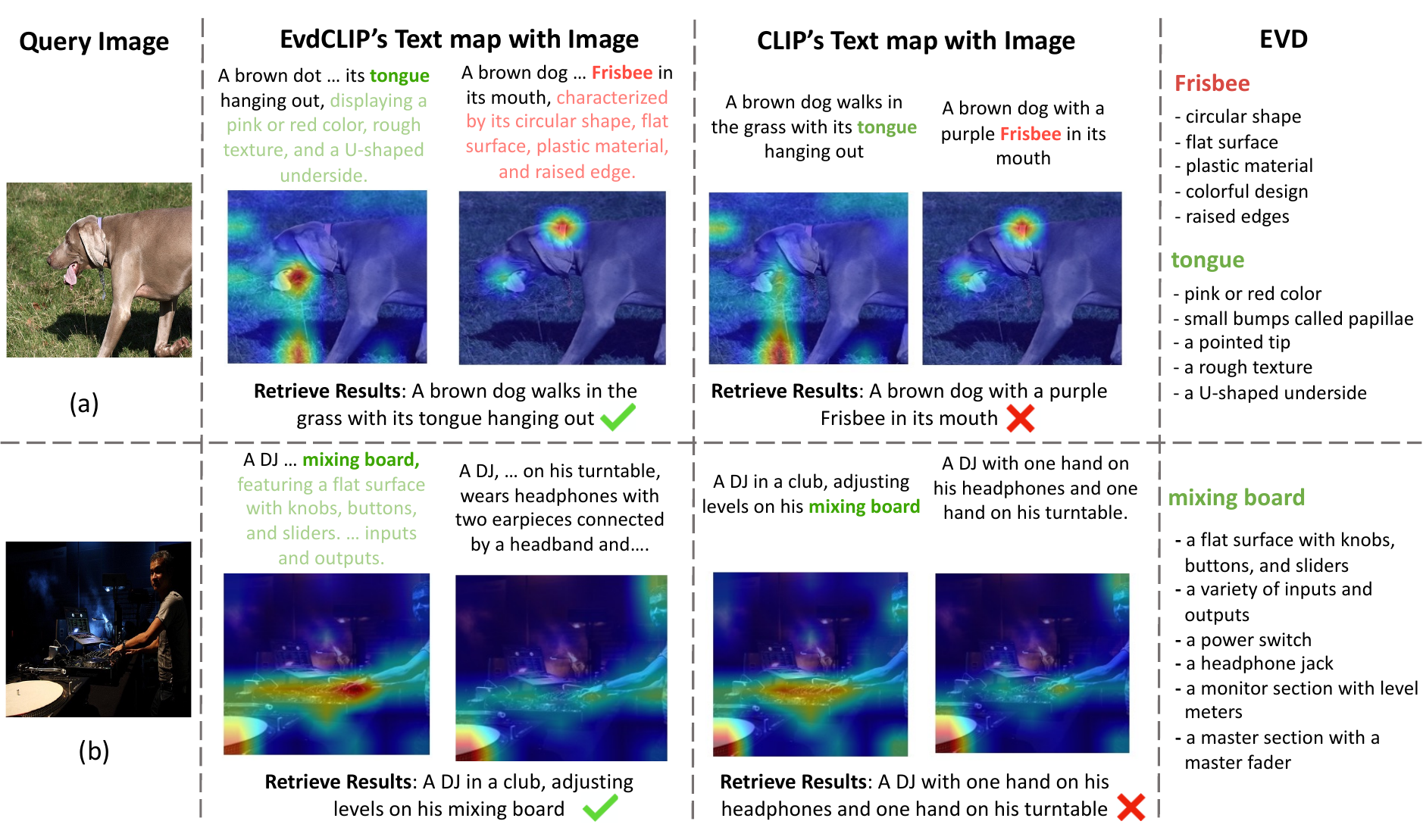}
  \caption{EvdCLIP focuses on significant regions of the image that are semantically related to the entity. Visualization examples of image-to-text retrieval are provided. We present image queries (the first column) along with four heatmaps.}
  \label{fig:vilization}
\end{figure}

\subsection{Ablation Studies}

% In our framework, the method of collecting high-quality visual entities is crucial, as the poor-quality entity set may introduce more distractions during image-text alignment.
% Considering the superior semantic understanding capabilities of large language models, we opted to utilize LLMs for extracting visual entities. The experimental results, displayed in Table~\ref{ablation-ecm}, demonstrate that our designed approach's performance improvement surpasses that of existing Parts of Speech Tagging (POS)\cite{banko2004part} and Named Entity Recognition (NER)\cite{souza2019portuguese} approaches. Specifically, as illustrated in Figure~\ref{fig:entity_nov}, the POS and NER methods may extract numerous non-visual terms that lack obvious visual features, such as "trick", "front", etc. Additionally, traditional NER methods struggle to extract generic nouns like "skateboard" effectively. Our approach excels in extracting visual terms more accurately through well-designed prompt templates and the powerful semantic capabilities of large language models.

\subsubsection{\textbf{Description Types}}\label{ab_destype}
Entity descriptions in our paper are of two types: conceptual descriptions from sources like WordNet~\cite{kilgarriff2000wordnet} and visual descriptions generated by our method. Table~\ref{ablation-4} compares the results of WordNetCLIP and EvdCLIP. WordNet provides only slight improvements in image-text retrieval, because its definitions are less relevant to visual understanding. In contrast, EvdCLIP’s visual descriptions better capture image content, leading to superior performance in cross-modal tasks.

% \begin{table}
%   \caption{Ablation studies of EVD-enhanced query generation methods. Vision encoder is ViT-B/32, Pre-Training dataset is Laion400M and the Fine-Tuning dataset is Flickr30k.}
%   % \caption{Ablation studies of different Evd strategies. The vision encoder is ViT-B/32, PT Data is NA and the fine-tuning dataset is Flickr30k. }
%   \label{ablation-3}
%   \centering
%   \adjustbox{width=1.0\linewidth}{
%   \begin{tabular}{c|lll|lll}
%      \hline
%      \multirow{2}{*}{EVD-enhanced Formulation}   &  \multicolumn{3}{c|}{I2T Retrieval} & 
%      \multicolumn{3}{c}{T2I Retrieval}  \\
%      &   R@1 & R@5 & R@10 & R@1 & R@5 & R@10 \\
%      \hline
%      CLIP &  64.8 & 85.7  & 49.2 & 79.3 & 49.2 & 79.3\\
%      \hline
%      DetCLIP &  65.3 & 87.0  & 50.3 & 79.8 & 49.2 & 79.3 \\ 
%      DesCLIP & 65.7 & 87.7  & 50.8 & 80.0 & 49.2 & 79.3 \\ 
%      CuPL & 66.3 & 88.0  & 51.5 & 80.2 & 49.2 & 79.3\\
%      CLIP-GPT& 66.1 & 88.2  & 51.3 & 80.1 & 49.2 & 79.3\\
%      LaBo & 66.1 & 88.2  & 51.3 & 80.1 & 49.2 & 79.3\\
%      \hline
%      EvdCLIP &  \textbf{66.9} & \textbf{88.6}  & \textbf{52.0} & \textbf{80.5} & 49.2 & 79.3 \\
%      \hline
%   \end{tabular}}
% \end{table}

\subsubsection{\textbf{Large Language Models}}
% Next, we conduct experiments using EvdCLIP equipped with various large language models, including ChatGPT, GPT-3, Llama-13B, and Vicuna-13B. As shown in Table~\ref{ablation-4}, GPT-3 and ChatGPT outperform others. Examples of description generated from various LLMs and a more detailed analysis are provided in Appendix~\ref{ab_llm}.
We test EvdCLIP with various LLMs, including ChatGPT, GPT-3, Llama-13B, and Vicuna-13B. As shown in Table~\ref{ablation-4}, GPT-3 and ChatGPT perform best. Examples and further analysis are in Appendix~\ref{ab_llm}.

\begin{figure}[t]
  \centering
  \includegraphics[width=0.5\textwidth]{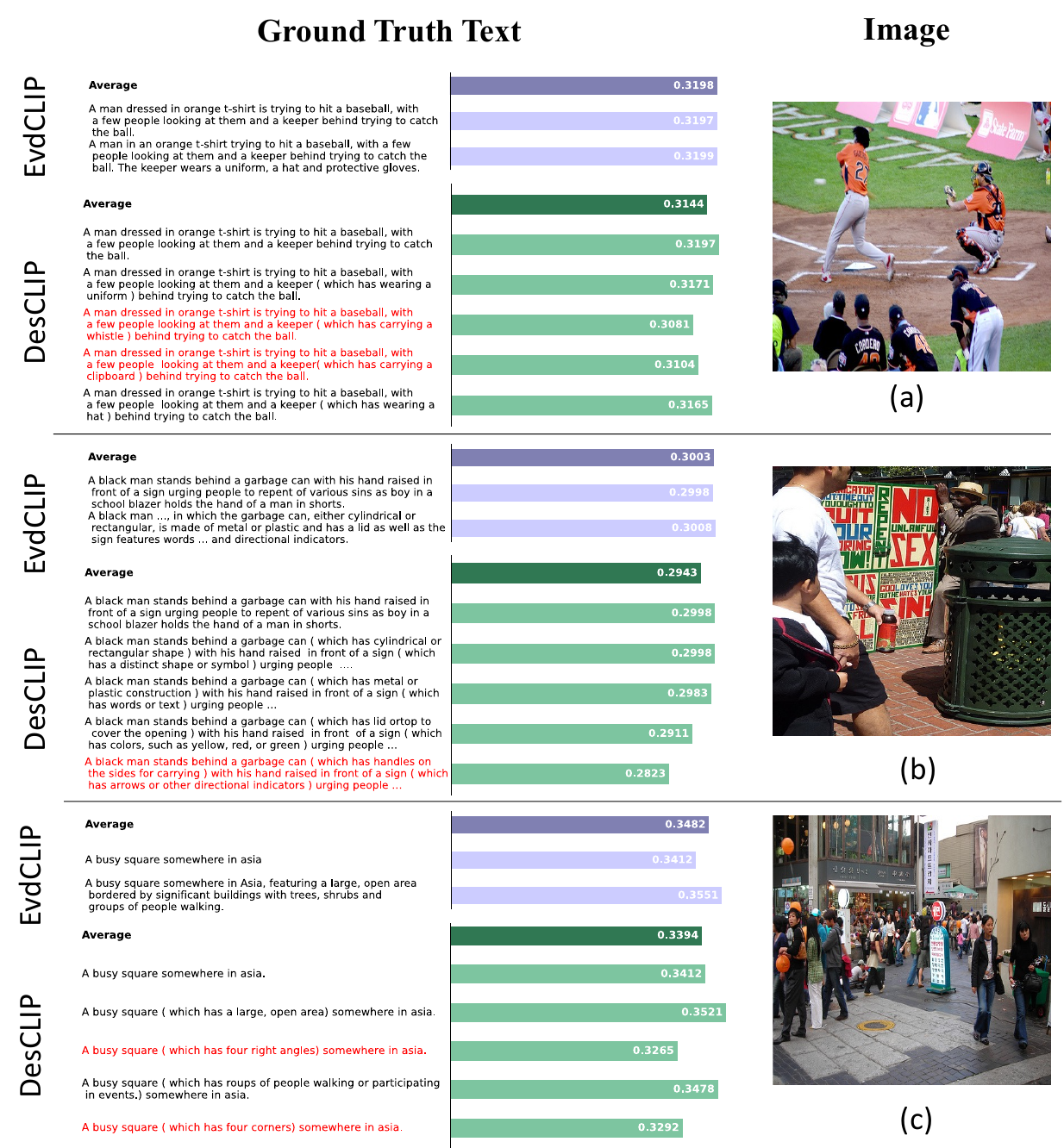}
  \caption{Comparison Between EvdCLIP and DesCLIP. The second column represents the image query. The first column shows the similar scores between Ground Truth and the image. The text in red annotates the errors.}
  \label{fig:fatal}
\end{figure}

\subsubsection{\textbf{EVD-enhanced Query Methods}}\label{experiment:evdmethod} 

We analyze the impact of different description enhancement methods. As shown in Table~\ref{ablation-overall}, DetCLIP adds concept descriptions for entities, resulting in only slight enhancement. 
DesCLIP adds visual descriptions, offering better performance than DetCLIP, but it suffers from noise and low-quality integration issues. 
CLIP-GPT and LaBo are designed for image classification denoising, but they fail to dynamically adjust for query content, limiting their performance. EvdCLIP outperforms all these methods. With EaRW and our training strategy, EvdCLIP efficiently filters and utilizes EVD based on the query, generating high-quality EVD-enhanced queries. 
% As shown in Table~\ref{ablation-3}, for PDF, where NA refers to treating multiple descriptions of entity equally, top$x$ refers to taking the EVD-text matching images with the top$x$ highest scores for training.
% EvdCLIP\_PDF (NA) is better than EvdCLIP\_CDF (NA), indicating that the parallel description can utilize visual descriptions more effectively. 
% Compared to CDF, PDF allows independent expression of each entity's visual description, enabling better capture of the clues provided by each description. 
% Additionally, PDF offers the potential for incorporating denoising methods to mitigate the impact of visual description noise.

% \subsubsection{Denosing Methods} 
% Finally, we investigate the impact of different denoising methods. As shown in Table~\ref{ablation-3}, for the PDF, both top$x$ and LCD outperform NA. This indicates that by filtering or correcting unreasonable visual descriptions, the impact of noisy samples on model training can be reduced.
% As illustrated in Figure~\ref{fig:noise_case}, unlike the rigid sample filtering of top$x$, LCD can adaptively modify labels to address noise problems, which is more flexible and retains more information. Thus LCD achieves better performance. 
\subsubsection{\textbf{Entity Collection Methods}}
In our framework, collecting high-quality visual entities is crucial. Ablation experiments on  entity collection methods can be seen in Appendix~\ref{ab_entity}.

\subsubsection{\textbf{The Number of Description $H$}}

 We also conduct ablation experiments to explore the effect of $H$ on model’s performance,  which can be found in Appendix~\ref{number_evd}.

\begin{figure}[t]
\centering
\includegraphics[width=1.\linewidth]{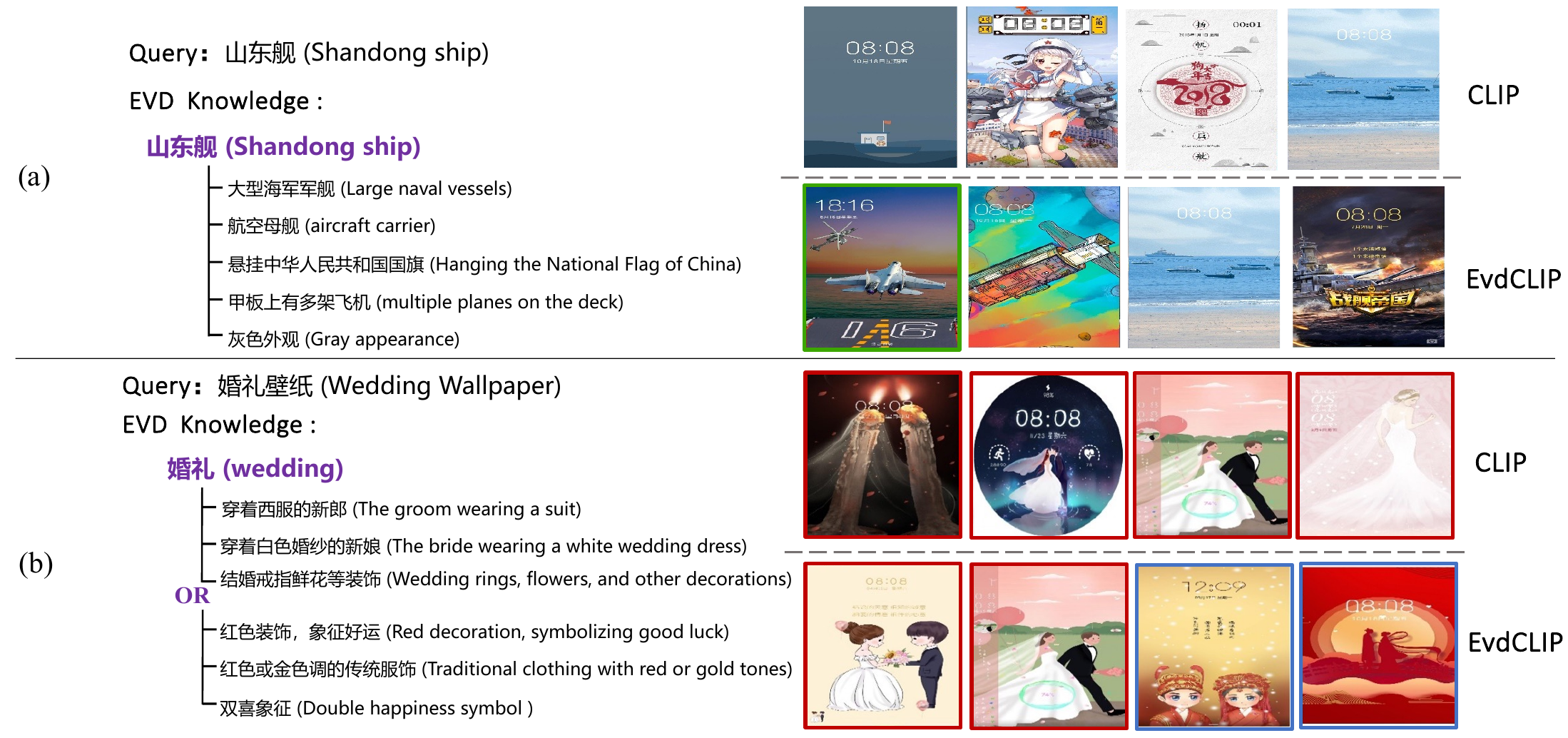}
\caption{Examples of Huawei Wallpaper Retrieval. The left is query and the right displays  top-4 retrieval results. (a) Images highlighted in green are user-satisfied; (b) Results highlighted in red depict Western weddings, while those in blue represent traditional Chinese weddings.}
\label{fig:edit_case}
\end{figure}

% \begin{figure*}
%   \includegraphics[width=\textwidth]{Figures/failure_case_paper.pdf}
%   \caption{Examples of failure case in EvdCLIP. The first column represents the image query. The second column shows the similar scores between Ground Truth and the image. The text in red annotates the errors. The third column displays the similar scores between the EvdCLIP's top prediction and the image.}
%   \label{fig:fatal}
% \end{figure*}

\subsection{Qualitative Analysis}

\subsubsection{\textbf{Superiority of EVD}}

 We use the Integrated Gradients algorithm~\cite{qi2019visualizing} to demonstrate how EVD helps the model focus on relevant image regions. In Figure~\ref{fig:vilization}(a), CLIP struggles to distinguish between the "frisbee" and the "tongue" in the dog's mouth, leading to inaccurate results. EVD enables EvdCLIP to differentiate these entities by emphasizing features like the "U-shaped underside" of a "tongue" versus the "circular shape" of a "frisbee." In Figure~\ref{fig:vilization}(b), CLIP struggles with the few-shot entity "mixing board", while EvdCLIP, guided by the visual description "a flat surface with knobs, buttons, and sliders," achieves better alignment. In summary, EVD helps EvdCLIP focus on semantically relevant regions, improving retrieval accuracy.

\subsubsection{\textbf{Superiority of EaRW}}
% Next, we conduct a qualitative analysis of EaRW's advantages. Due to LLM's hallucination, some entities may be misunderstood. For example, in Figure~\ref{fig:fatal} (a), "carrying a whistle" and "carrying a clipboard" are incorrect descriptions of the entity "keeper," leading to inaccurate retrieval results. EaRW, trained on specified datasets $D_{EQR}$, recognizes these descriptions as intrusive and filters them out during query rewriting.
% Beyond hallucination, EaRW also mitigate EVD's noise. In Figure~\ref{fig:fatal}(b), the description "handles on the side" does not match the "garbage can" in the image. EaRW, learning the appearance preferences of high-frequency entities, selectively includes relevant descriptions in the query.
% EaRW is also effective in resolving entity ambiguity. As shown in Figure~\ref{fig:fatal}(c), the "square" can refer to either plaza features or geometric features. Geometric descriptions could reduce matching accuracy, but EaRW adapts by selecting "plaza"-related descriptions to enhance the query based on query's context.
% In summary, EaRW effectively mitigates the challenges of hallucinations, noise, and ambiguity in EVD.

% 我们接下来定性分析下EaRW的优势，如图所示，第一行的示例中，
We then qualitatively analyze the advantages of EaRW. Due to LLM-induced hallucinations, some entities may be misinterpreted. For example, in Figure~\ref{fig:fatal}(a), "carrying a whistle" and "carrying a clipboard" are incorrect descriptions for the entity "keeper," resulting in inaccurate retrievals. EaRW, trained on the dataset $D_{EQR}$, identifies these intrusive descriptions and filters them out during query rewriting.
Beyond hallucinations, EaRW also reduces noise. In Figure~\ref{fig:fatal}(b), the description "handles on the side" does not match the "garbage can" in the image. EaRW learns the appearance preferences of high-frequency entities and selectively incorporates relevant descriptions into the query.
EaRW also effectively resolves entity ambiguity. As shown in Figure~\ref{fig:fatal}(c), the term "square" can refer to either plaza features or geometric shapes. Geometric descriptions may reduce matching accuracy. EaRW adapts by choosing "plaza"-related descriptions based on the query's context. In summary, EaRW effectively mitigates the challenges of hallucinations, noise, and ambiguity in EVD.

\subsection{Methodological Editability}
Unlike black-box models, our framework demonstrates editability through the incorporation of EVD.

\subsubsection{\textbf{Novel Knowledge Injection:}} CLIP is limited to understanding concepts that existed before its training. In contrast, our approach enables the model to grasp novel concepts by integrating visual descriptions. For instance, in Huawei Wallpaper Retrieval, as shown in Figure~\ref{fig:edit_case} (a), when the query \begin{CJK*}{UTF8}{gbsn} "山东舰" \end{CJK*} is used, the CLIP model produces poor retrieval results. By constructing appropriate descriptors, our model can recognize that \begin{CJK*}{UTF8}{gbsn}  "山东舰" \end{CJK*}  refers to an aircraft carrier and retrieve images that satisfy the user’s intent.

% \subsubsection{\textbf{Entity Bias Correction}} Visual descriptions enable manual correction to address biases in recognition systems. Since our model's decisions rely on human-readable text, any alteration to visual descriptions will modify the decision-making process. Figure~\ref{fig:edit_case} (b) illustrates how bias issues can be addressed through the editing of EVD. When querying \begin{CJK*}{UTF8}{gbsn} 
% "婚礼壁纸" % 中文文本
% \end{CJK*}, the CLIP method exhibits biases. Specifically speaking, when dealing with concepts  \begin{CJK*}{UTF8}{gbsn} 
% "婚礼" % 中文文本
% \end{CJK*}, CLIP's retrieval results may exhibit a bias towards Western weddings, potentially due to the model fitting more data from Western cultures during training. We can manually construct wedding descriptions encompassing Chinese traditional weddings, effectively guiding the model to explore the diversity of concepts rather than relying solely on biased pretraining knowledge. 
% More details of Editability can be fonud in Appendix~\ref{app:edit}.

\subsubsection{\textbf{Entity Bias Correction:}} EVD allow for manual bias correction in recognition systems. Since EvdCLIP's decision relies on EVD, altering descriptions will impact outcomes. Figure~\ref{fig:edit_case} (b) shows how editing EVD can address bias. For instance, when querying \begin{CJK*}{UTF8}{gbsn} "婚礼壁纸" \end{CJK*}, CLIP may favor Western weddings due to biased training data. By incorporating descriptions of traditional Chinese weddings, we guide the model to explore a more diverse range of concepts. More details on editability are provided in Appendix ~\ref{app:edit}.

\section{Conclusion}
In this paper, we propose EvdCLIP, which employs entity visual descriptions generated by LLMs as auxiliary information to guide visual-textual alignment. To address the noise and low-quality issue of EVD integration, we develop an EVD-aware Rewriter, which utilizes EVD knowledge and the generative capabilities of pretrained language models to rewrite query elegantly. Our approach achieves performance improvements across various vision-language retrieval benchmarks with multiple benefits, including enhanced model generalizability, transferability, and editability. EvdCLIP has several limitations that also point to the directions for our future optimizations. These issues are discussed in Appendix ~\ref{appen_limit}.

% \section{Acknowledgements}
% This work is supported by the Major Key Project of PCL under grant No. PCL2023A06, the National Key Research and Development Program of China under grant No. 2022YFB3105000, and the Shenzhen Key Lab of Software Defined Networking under grant No. ZDSYS20140509172959989.

\clearpage
\bibliographystyle{aaai25}
\bibliography{aaai25}

\clearpage

\appendix

\section{DATASETS}
\subsection{EVD-enhanced Query Rewriting Dataset Construction}\label{warmup_instruct}
\begin{figure}[ht]
\centering
\includegraphics[width=\linewidth]{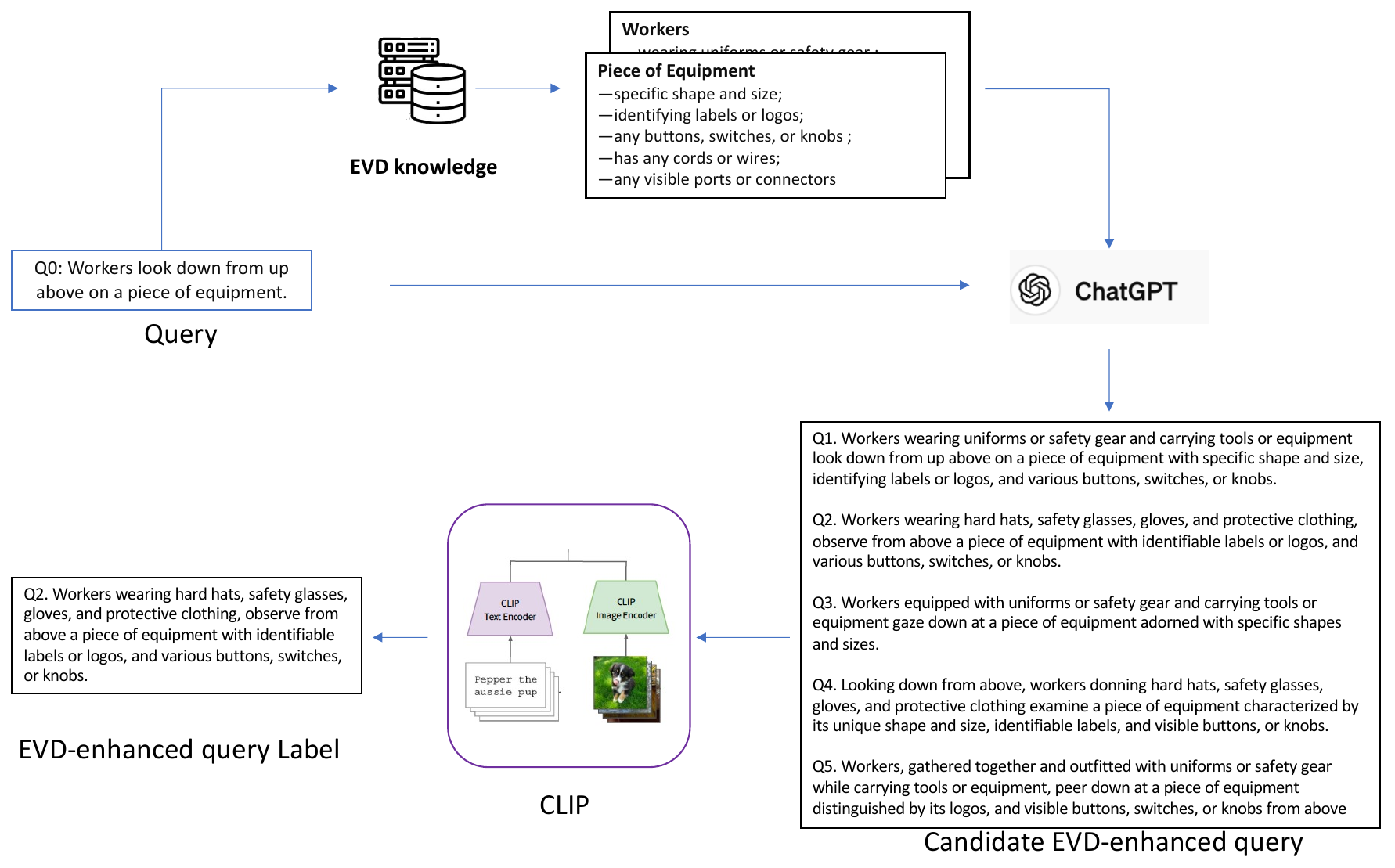}
\caption{Illustration of the warmup dataset construction pipeline.}
\label{fig:warmupdata_case}
\end{figure}

Our warm-up dataset is constructed as shown in Figure~\ref{fig:warmupdata_case}. LLMs serve as teachers for the EaRW.  We generate multiple ($k$) candidate EVD-enhanced query labels for each original query by using prompt templates in the Figure~\ref{fig:prompt}. These candidate sets vary in quality. Considering that high-quality and low-noise EVD-enhanced queries enhance the CLIP model's retrieval accuracy, we input them together with the original query into CLIP and calculate the matching score with the corresponding images. Samples that score higher than the original query and are ranked first are selected for the EVD-enhanced query rewriting dataset $D_{EQR}$.

More specifically, we use GPT-3.5 as the teacher model to generate multiple candidate EVD-enhanced queries for each original query. We then employ a pre-trained CLIP model to evaluate and select the most effective EVD-enhanced queries to be included as rewrite query labels for the original query. For each original query, we collect $k$ EVD-enhanced query labels. In this paper, $k$ is set to 5. The final $D_{EQR}$ is represented as a tuple $(x: \{y_i, s_i\}_{i=1}^k)$, where $x$ is the original query, $y_i$ is the $i$-th EVD-enhanced query label for $x$, $s_i$ is the corresponding score, and $k$ is the number of pseudo-labels for each $x$. We create the training set of $D_{EQR}$ using the training sets of benchmark datasets such as Flickr30k or MSCOCO, and similarly, we create the test set of $D_{EQR}$ using their test sets.

By leveraging the powerful contextual understanding of large language models and the feedback from CLIP, the rewriting query labels collected through our method effectively capture the contextual preferences of EVDs and the entity preferences within the dataset.

\subsection{Huawei Business Datasets}\label{huawei data}
This dataset is collected from Huawei Mobile Scene Search Service and comprises a significant number of Chinese image-text pairs. The datasets are categorized into four types: 
(1) Theme dataset consists of image-text pairs related to mobile phone themes.
(2) Wallpaper dataset consists of image-text pairs representing mobile phone wallpapers.
(3) Lock-Screen dataset contains image-text pairs designed for mobile phone lock screens.
(4) Icon dataset is collected from image-text pairs representing mobile phone icons.
Table~\ref{ablation-dataset} provides their statistics, including sample sizes for the train/validation/test, 
as well as the average token length and the average number of entities in text queries

\begin{table}

  \caption{Statistics of Huawei business datasets. Avg. TextLen and Avg. EntNum refer to the average token length and the average number of entities in a text query, respectively.
  }
  \label{ablation-dataset}
  \centering
  \resizebox{\linewidth}{!}{  
  \begin{tabular}{c|c|c|c|c|c}
     \hline
     Dataset & \#Train & \#Val & \#Test & \#Avg. TextLen & \#Avg. EntNum \\
     \hline
     Theme & 144k  & 18k & 18k  & 5.6  &  1.1  \\
     \hline
     Wallpaper & 56k & 7k & 7k  & 7.3  & 1.8  \\
     \hline
     Lock-Screen  & 48k & 6k & 6k  & 6.5  &  1.3  \\
     \hline
     Icons  & 32k & 4k  & 4k  & 6.2  &  1.2 \\
     \hline
  \end{tabular}}
\end{table}

\section{Prompt Template}\label{appendix_prompt}
\begin{figure}[t]
\centering
\includegraphics[width=1.0\linewidth]{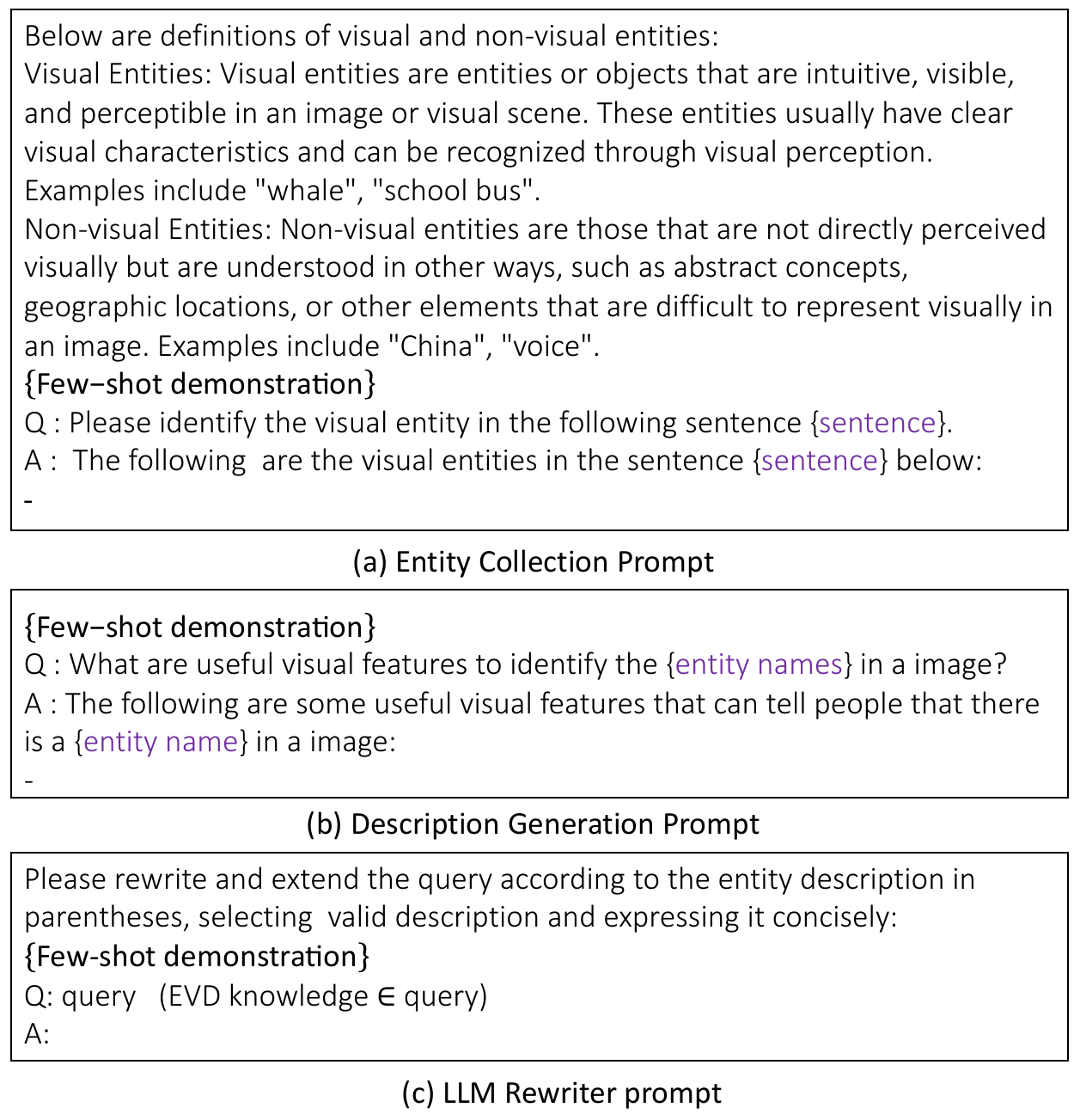}
\caption{The prompt templates used for the LLMs. $\left\{ \text{Few-shot Demonstrations} \right\}$ denotes 1-3 instance.}
\label{fig:prompt}
\end{figure}

The prompt template used for the LLMs in our paper is shown in Figure~\ref{fig:prompt}.
Our prompt follows the formulation of [instruction, demonstrations, input], where the input is a question. The instruction is straightforward and the demonstrations are 1-3 random examples from training sets and are kept constant across all runs, mainly to illustrate the task-specific output format illustration.

\begin{figure}[t]
\centering
\includegraphics[width=\linewidth]{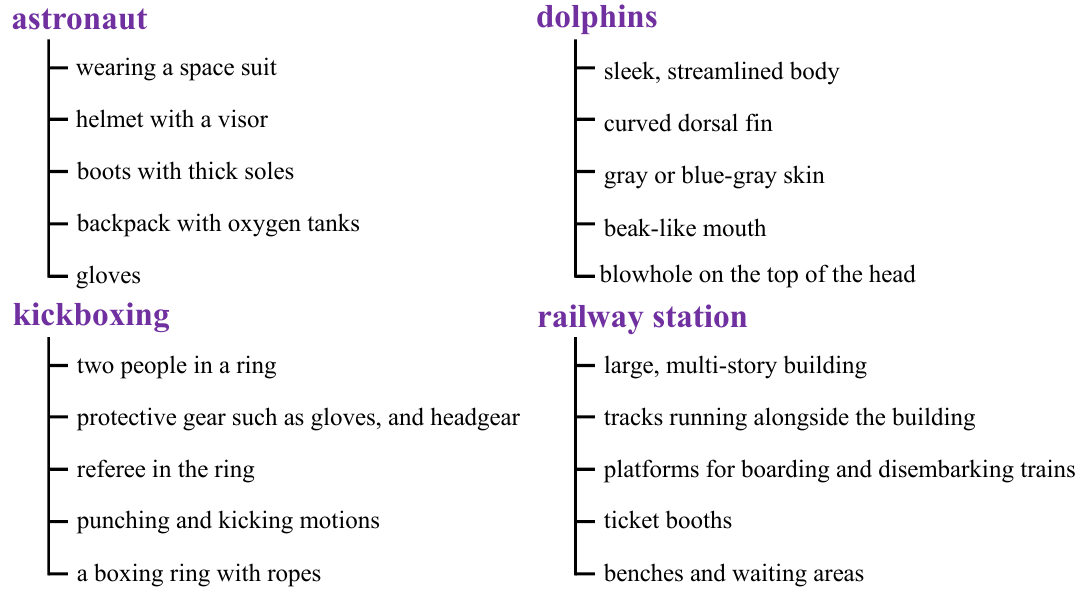}
\caption{Examples of entity visual descriptions.}
\label{fig:evd_case}
\end{figure}

\begin{figure}[t]
\centering
\includegraphics[width=1.0\linewidth]{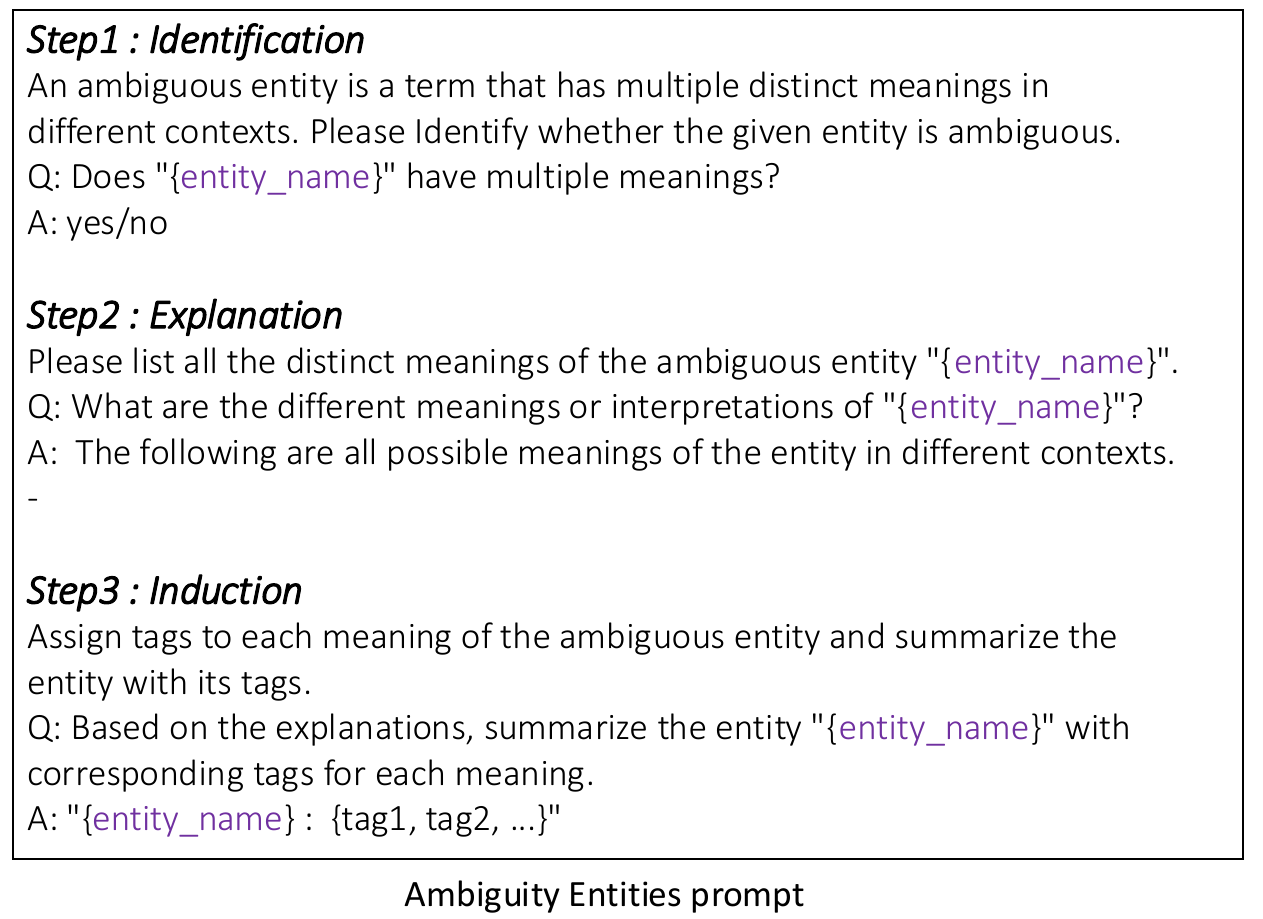}
\caption{The prompt templates used for ambiguous entity.}
\label{fig:aeae_prompt}
\end{figure}

\section{Example of EVD}\label{app:exaple_evd}
In Figure~\ref{fig:evd_case}, the visual descriptions showcase the remarkable ability of Large Language Models to capture and articulate distinct visual attributes of various entities. These descriptions go beyond basic elements such as colour and shape, extending to specific object parts and functionalities that enhance recognition and distinction in visual scenes.

For example, the description of an astronaut highlights specialised equipment such as a visor, thick-soled boots, and an oxygen-carrying backpack, emphasizing the model's capability to identify functionally relevant features. Similarly, the depiction of dolphins, focusing on sleek bodies, dorsal fins, and characteristic beak-like mouths, captures the essential features that visually distinguish these marine animals.

In sports scenes such as kickboxing, the descriptions detail the environment and action, including the presence of two fighters in protective gear, a referee, and the dynamics of punching and kicking within a roped ring, providing a vivid, context-rich snapshot. These description encompasses structural and functional aspects, such as multi-storey buildings, adjacent tracks, platforms, ticket booths, and waiting areas, providing a comprehensive picture of the bustling activity typical of such environments.

These rich, detailed descriptions demonstrate the model's strength in generating diverse, accurate visual representations from textual cues. This ability not only helps to create more engaging content but also enhances the model's utility in applications requiring detailed visual understanding and interpretation.

\section{Copying with  Ambiguity  Entities}\label{app:multi-mean}

In this section, we discuss how to handle ambiguous entities with multiple meanings within the EvdCLIP. Such entities, like "bank," which can refer to either a financial institution or a riverbank, present challenges for model inference and training. To address this issue without significantly increasing inference and training costs, we propose an "identification-explanation-induction" approach that adapts the construction of the entity set to handle ambiguous entities.

The prompt in shown in Figure~\ref{fig:aeae_prompt} and the specific steps are as follows: \textbf{(1) Identification Ambiguous Entities:} First, we use large language models (LLMs) to automatically identify entities with multiple meanings.
\textbf{(2) Explanation Ambiguous Entities:} After the ambiguous entity is identified, the language model generates explanations for all possible meanings of the entity. For example, for "bank" the model would generate two explanations: one referring to a financial institution and the other to a riverbank.
\textbf{(3) Inductive Storage of Ambiguous Entities:} In this step, we assign a specific tag to each meaning and add it to the entity set. In this way, each ambiguous entity in the set has multiple tags representing its different meanings. For example, the entity "bank" would be tagged as "bank: {financial institution, riverbank}".

During the subsequent generation of EVDs, the ambiguous entity simply needs to iterate through its tags, replacing the \{entity name\} in Figure~\ref{fig:prompt} with \{entity name + tag name\} to generate descriptions. This approach avoids the need to modify the model structure or add new modules during subsequent training and inference, thereby maintaining the efficiency and simplicity of the inference process.

As shown in the Figure~\ref{fig:fatal}, existing methods randomly select descriptions or use fixed weights for denoising, which are ineffective at contextually filtering noise and thus struggle to handle ambiguous entities. In contrast, our EaRW can select appropriate descriptions based on the context, and thus construct high-quality EVD-enhanced queries. 

\section{Implantation Detail}\label{appendix_detail}
In our experiments, the vision encoder comprises two variants: the Vision Transformer (ViT-B/32, ViT-B/16 and ViT-L/14) and ResNet (RN50 and RN101), while the text encoder is a vanilla Transformer \cite{vaswani2017attention}  following CLIP. The input image is resized to 224 × 224 and the input  sequence is truncated or padded to 77. 
In the construction of the EVD, we utilize ChatGPT to gather entities from the training sets of the Flickr30k and MSCOCO datasets. After collecting entities, we filter out low-frequency entities to ensure the relevance and robustness of the dataset. This process result in the collection of approximately 10k entities ($M= 10237$). Subsequently, we employ ChatGPT to generate visual descriptions for these entities. As illustrated in Figure~\ref{fig:prompt}, we do not impose a limit on the number of descriptions generated for each entity during the initial phase. However, for consistency in our subsequent EVD enhancement and ablation experiments, we fix the number of extracted descriptions $H=5$ for each entity. In the Huawei business dataset, we use the PanGu large language model insted of ChatGPT for our experiments. 

EaRW is initialized using the pre-trained T5-large model (770M parameters), making it more feasible for real-world deployment. We conduct the warm-up phase of EaRW with a learning rate of 3e-5, a batch size of 8, and over 20 epochs. For the Rank Preference Optimisation (RPO) model, we set the learning rate to 5e-7, with a batch size of 16, across 5 epochs, and used a rank length of 5. The weight of the SFT loss $\beta$ is set to 0.2, and the probability of random rewriting during CLIP fine-tuning $p$ is set to 0.6.

We build EvdCLIP based on fine-tuning on pre-trained CLIP model~\cite{radford2021learning}. For the hyper-parameters used for fine-tuning CLIP, we employ the Adam optimizer \cite{kingma2014adam} with weight decay of 1e-3 and batch size is set to 256. The total number of fine-tuning epochs is set to 20. The initial learning rate is set to 1e-6 and a cosine learning rate decay scheduler is applied. We apply a warm-up strategy for the initial 2k steps. Following previous work~\cite{radford2021learning}, we use recall R@$h (h = 1, 5, 10)$ as the evaluation metrics.

In the VLR framework, we implemented various description enhancement methods, including DetCLIP~\cite{yao2022detclip}, DesCLIP~\cite{menon2022visual}, CLIP-GPT~\cite{maniparambil2023enhancing}, and LaBo~\cite{yang2023language}, with the following details:
\begin{itemize}
    \item DetCLIP~\cite{yao2022detclip}: Generates five parallel text queries, including the original query and four with descriptions randomly selected from WordNet.
    \item DesCLIP~\cite{menon2022visual}: Similar to DetCLIP but uses descriptions from the EVD dictionary shared with EvdCLIP.
    \item CLIP-GPT~\cite{maniparambil2023enhancing}: Builds on DesCLIP by adding a self-attention adapter after the text tower to enhance noise filtering.
    \item LaBo~\cite{yang2023language}: Filters descriptions using a combined distinctiveness and coverage score. Distinctiveness measures a description’s effectiveness in distinguishing entities, while coverage evaluates its contribution to the diversity of the description set. Descriptions are ranked by their combined scores, and the top five are selected. Since scores can vary across query contexts, we aggregate scores across all relevant queries before filtering. A learnable weight matrix is applied to optimize the selected descriptions.
\end{itemize}

CLIP-GPT and LaBo are originally designed as descriptive denoising strategies for image classification, thus they do not dynamically filter and integrate EVDs based on query content.
This comparative implementation allows us to assess the effectiveness of these methods in improving vision language retrieval and to understand the specific contributions of each approach.

\section{Ablation Study}

\subsection{Large Language models}\label{ab_llm}
\begin{figure}[t]
\centering
\includegraphics[width=0.99\linewidth]{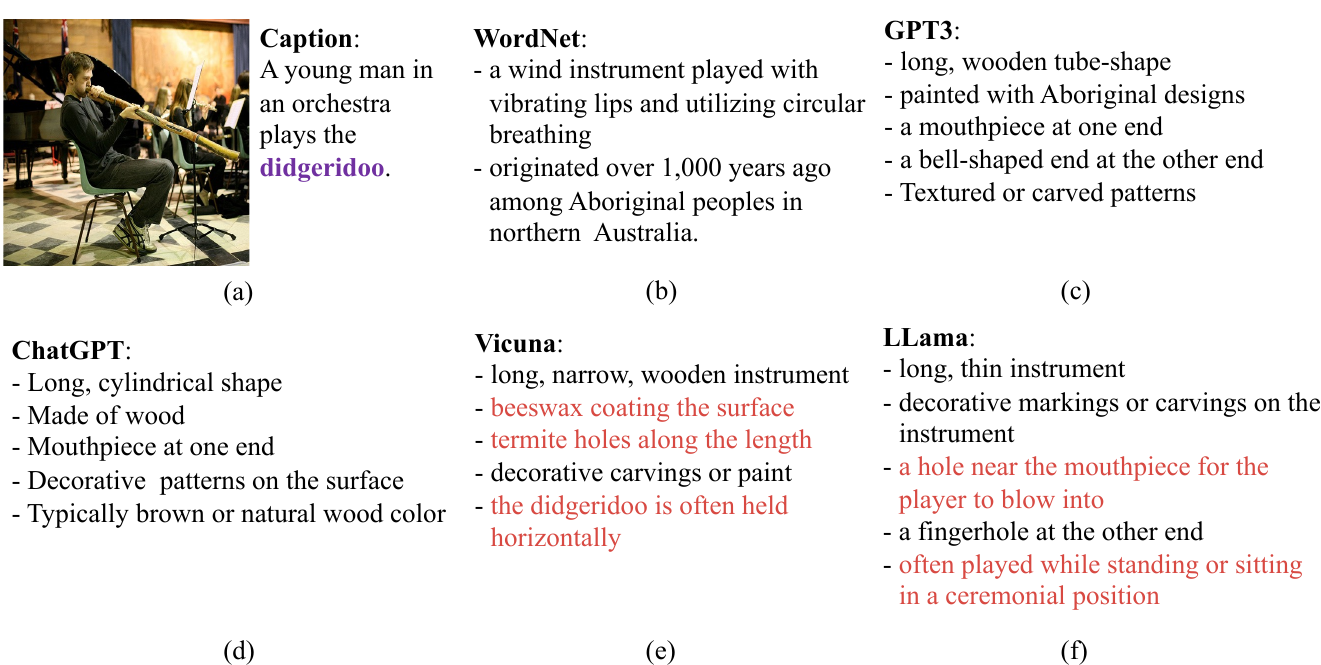}
\caption{Example of entity description from various sources. (a) Image and the corresponding caption. (b)-(f) are descriptions of the "didgeridoo", where (b) is the conceptual description from WordNet, and (c)-(f) are visual descriptions from LLMs, descriptions in red are disruptive descriptions.}
\label{fig:des_source}
\end{figure}

As shown in Table~\ref{ablation-4}, experimental results reveal that EvdCLIP, equipped with any LLMs, can generate visually helpful descriptions for the model.
Different LLMs show slight variations in performance improvement. GPT-3 and ChatGPT outperform others. The examples of description generated from various LLMs are provided in Figure~\ref{fig:des_source}.

We speculate that under our prompt template, compared with ChatGPT and GPT-3, Vicuna and Llama have two defects: On the one hand GPT-3 and ChatGPT generate more concise descriptions that capture important visual features of the entity, while Vicuna and Llama generate overly detailed and abstract descriptions that introduce more distractions such as "beeswax coating the surface " and "often played while standing or sitting ..."; On the other hand, Vicuna and Llama would have more illusions, such as "the didgeridoo is often held horizontally". Compared to ChatGPT, although GPT-3 may have shortcomings in complex logical reasoning, such ability is not critical for generating EVD.

\subsection{Entity Collection Methods}\label{ab_entity}
\begin{figure}[t]
    \centering
    \includegraphics[width=0.98\linewidth]{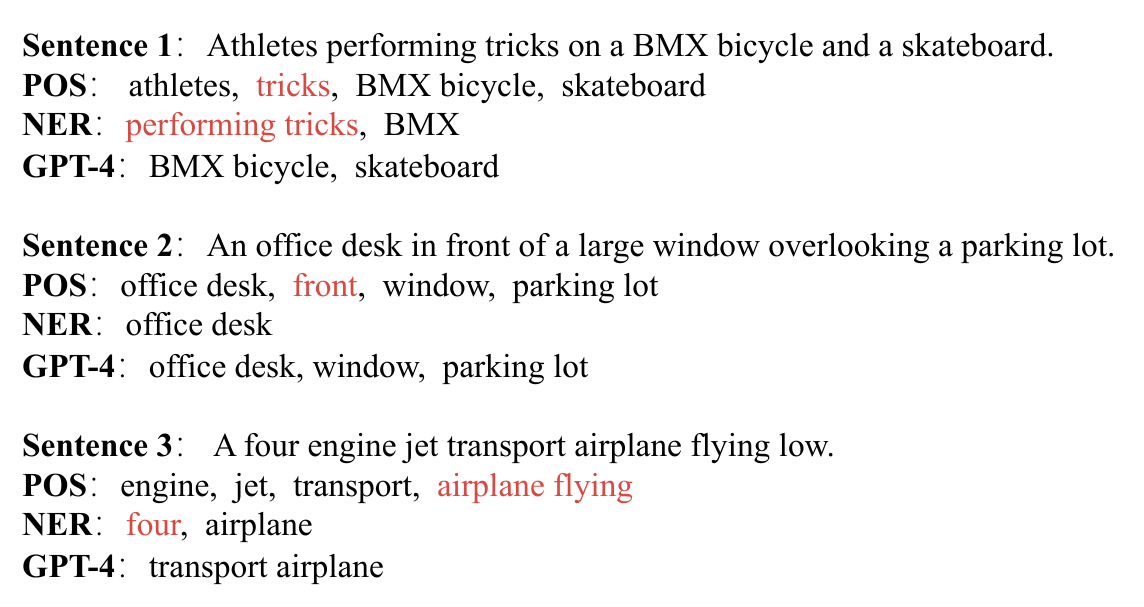}
    \caption{Comparison of different entity collection methods, where red-marked entities denote unreasonable visual items.}
    \label{fig:entity_nov}
\end{figure}

In our framework, the method of collecting high-quality visual entities is crucial, as the poor-quality entity set may introduce more distractions during image-text alignment.
Considering the superior semantic understanding capabilities of large language models, we opted to utilize LLMs for extracting visual entities. The experimental results, displayed in Table~\ref{ablation-ecm}, demonstrate that our designed approach's performance improvement surpasses that of existing Parts of Speech Tagging (POS)~\cite{banko2004part} and Named Entity Recognition (NER)~\cite{souza2019portuguese} approaches. Specifically, as illustrated in Figure~\ref{fig:entity_nov}, the POS and NER methods may extract numerous non-visual terms that lack obvious visual features, such as "trick", "front", etc. Additionally, traditional NER methods struggle to extract generic nouns like "skateboard" effectively. Our approach excels in extracting visual terms more accurately through well-designed prompt templates and the powerful semantic capabilities of large language models.

\begin{table}
  \caption{Ablation studies of entity collection methods.
  The vision encoder is ViT-B/32, Pre-Training dataset is Laion400M and Fine-tuning dataset is Flickr30k.}
  \label{ablation-ecm}
  \centering
\adjustbox{width=1\linewidth}{
  \begin{tabular}{l|ll|ll}
     \hline
     \multirow{2}{*}{Methods} & \multicolumn{2}{c|}{I2T Retrieval} & \multicolumn{2}{c}{T2I Retrieval}  \\
     & R@1 & R@5  & R@1 & R@5  \\
     \hline
     Part of Speech Tagging & 90.2 & 98.6  & 75.2 & 93.0  \\
     \hline
     Named Entity Recognition  & 89.8 & 98.3 & 74.9 & 92.8  \\
     \hline
     \multirow{1}{*}{LLMs (ChatGPT) based method} & \textbf{90.6} & \textbf{99.0}  & \textbf{75.5} & \textbf{93.4}  \\    
     \hline
  \end{tabular}}
\end{table}

\begin{figure}[ht]
\centering
\includegraphics[width=0.7\linewidth]{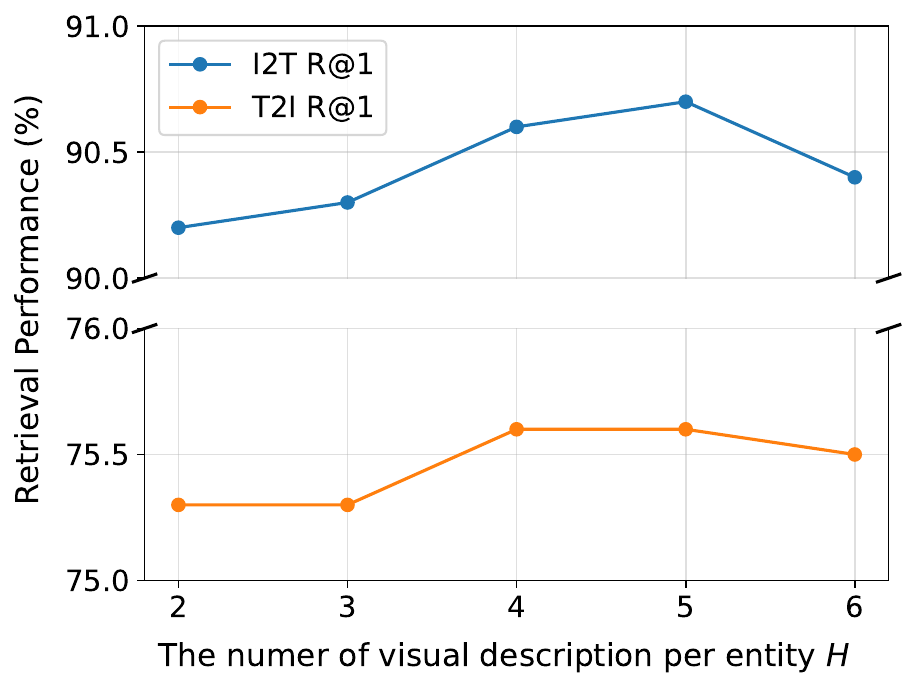}
\caption{Ablation study of the number of descriptions per entity $H$.}
\label{fig:number_case}
\end{figure}

\subsection{The number of descriptions per entity $H$}\label{number_evd}
We performe ablation experiments to examine the effect of varying the number of descriptions, denoted as $H$, on the model's performance. As shown in Figure~\ref{fig:number_case}, the model achieves optimal performance when $H=5$. If $H$ is too small, the diversity of the EVD may be insufficient, limiting the model's ability to capture nuanced visual concepts. On the other hand, increasing $H$ too much may introduce a higher number of low-quality or irrelevant descriptions, which could dilute the effectiveness of the model and potentially lead to performance degradation. This finding highlights the importance of balancing diversity and quality in the EVD to maximize the model's performance.

\section{Details of Methodological Editability}\label{app:edit}
Let us delve into the details of model editability. As discussed in this paper, editability allows users to modify the EVDs to influence the model's decision-making process. For knowledge injection, users can update the EVDs. During query rewriting, the newly injected knowledge is integrated to guide retrieval. To address bias issues, we add parallel definitions for entities, generating multiple parallel queries during the rewriting process. The final score is then computed by averaging the scores of these parallel queries. Similarly, our method can improve visual diversity by injecting different visual knowledge.

Specifically, for the term "wedding," we use large language models to generate different descriptors for "Western wedding" and "Chinese traditional wedding." During retrieval, multiple descriptors are randomly selected to create various wedding queries that reflect different descriptive preferences. By taking these different descriptions into account and averaging their scores, the retrieval results show a more diversified ranking that takes into account the diversity of the concept, effectively eliminating potential biases.
% \begin{figure}[t]
%   \centering
%   \includegraphics[width=\linewidth]{Figures/limit_apr.pdf}
%   \caption{EvdCLIP's Failure Case. The First column represents the image query. The second column shows similar scores between Ground Truth/Top prediction and  image.}
%   \label{fig:limit}
% \end{figure}
\

\section{Limitations}\label{appen_limit}
We further discuss the limitations of our approach. In addition to the description noise primarily addressed in this text, our method has several limitations:
(1) Dependence on the quality of the EVD dictionary: The effectiveness of our method heavily relies on the quality of the EVD, which serves as the basis of our approach. Future work will focus on improving the quality of the EVD dictionary to enhance the effectiveness of our method. 
(2) Interference from multiple entities: In VLR, a query often contains multiple entities, each with varying levels of importance in text-image matching. Over-describing for less important entities can confuse the model and degrade performance. Therefore, entity-awareness and the efficient extraction of key entities for description will be crucial in future work. (3) Query length issue: Some queries involve a large number of entities and descriptions, which will exceed CLIP's 77-token limit during query expansion. Addressing this challenge will be important for our future work.

\end{document}